\documentclass[11pt, a4paper]{article}
\usepackage{subfigure}
\usepackage[all]{xypic}

\usepackage{color, colortbl}
\usepackage[a4paper,margin=2.5cm]{geometry}

\newcommand{\R}{\mathbb{R}} 
\newcommand{\N}{\mathbb{N}}
\newcommand{\Hc}{\mathcal{H}}

\definecolor{Gray}{gray}{0.9}
\definecolor{LightCyan}{rgb}{0.88,1,1}

\usepackage{graphicx,
    amssymb,
    amsmath,
    amsthm,
    dsfont, 
    xcolor,
    mathtools,
    authblk,
    enumitem,
    tikz,
    mdframed, 
    todonotes,
}
 \usepackage{soul}
\usepackage{subeqnarray}
\usepackage{amsmath, amsfonts} 
\usepackage{algorithm,algorithmic}
\usepackage{graphicx}

\def\[{\begin{equation} }
\def\]{\end{equation} }

\newcommand{\RR}{\mathbb{R}}

\newtheorem{definition}{Definition}[section]
\newtheorem{proposition}{Proposition}[section]

\newtheorem{theorem}{Theorem}[section]

\usepackage[utf8]{inputenc}
\usepackage
{
    graphicx,
    amssymb,
    amsmath,
    amsthm,
    dsfont, 
    xcolor,
    mathtools,
    authblk,
    enumitem,
    tikz,
    mdframed, 
}
\usepackage[utf8]{inputenc}

\usepackage[pdffitwindow=false,
            plainpages=false,
            pdfpagelabels=true,
            pdfpagemode=UseOutlines,
            pdfpagelayout=SinglePage,
            bookmarks=false,
            colorlinks=true,
            hyperfootnotes=false,
            linkcolor=blue,
            urlcolor=blue!30!black,
            citecolor=green!50!black]{hyperref}

\usepackage[bf,normal]{caption}
\usepackage{graphicx}
\usepackage{mathptmx}       
\usepackage{helvet}         
\usepackage{courier}        
\usepackage{type1cm}        
\usepackage{makeidx}         
\usepackage{graphicx}        
\usepackage{multicol}        
\usepackage[bottom]{footmisc}
\usepackage{tikz}
\usetikzlibrary{patterns,intersections}
 \def\[{\begin{equation}}
\def\]{\end{equation}}
\usepackage{pgfplots}
\usepackage{pgfplotstable}
\pgfplotsset{compat=1.14}
\usepgfplotslibrary{groupplots}

\makeindex             

\begin{document} 

\title{Learning dynamical systems from data: \\ 
a simple cross-validation perspective}

\author[1]{Boumediene Hamzi}

\author[2]{Houman Owhadi}
 
\affil[1]{Department of Mathematics, Imperial College London, United Kingdom.  email: boumediene.hamzi@gmail.com}
\affil[2]{Department of Computing and Mathematical Sciences, Caltech, CA, USA. email: owhadi@caltech.edu}

 
%
%
\maketitle

\abstract{Regressing the vector field of a dynamical system from a finite number of observed states is a natural way to learn surrogate models for such systems. We present variants of cross-validation (Kernel Flows \cite{Owhadi19} and its variants based on Maximum Mean Discrepancy and Lyapunov exponents) as simple approaches for learning the kernel used in these emulators. 
}


\section{Introduction}

Linear stochastic models (autoregressive (AR), moving average (MA), ARMA models) and chaotic dynamical systems are natural predictive models for time series \cite{boxjen76,abarbanel2012analysis,kantz97,nielsen2019practical,shumway2010time}.

The prediction of chaotic systems from time-series (initially investigated in \cite{CASDAGLI1989}) has been investigated from the regression perspectives of support vector machines \cite{muller_svm,sayan_svm}, reservoir computing \cite{jaideep1,ott}, deep feed-forward artificial neural networks (ANN), and recurrent neural networks with long short-term memory (RNN- LSTM) \cite{Sindy,  bookmezic,havok,NIPS2003_2516}. Reservoir computing was observed to be  efficient for predictions but not very accurate for estimating Lyapunov exponents. On the other hand,  RNN-LSTM were observed to be accurate for estimating Lyapunov exponents but  not as good as reservoir computing for predictions (see \cite{survey_kf_ann} for a survey). 
Although Reproducing Kernel Hilbert Spaces (RKHS) \cite{CuckerandSmale} have provided strong mathematical foundations for analyzing dynamical systems \cite{BH4,BH3,BH6,BH10,BHSIAM2017,lyap_bh,bh2020a, bh2020b,klus2020data,KNPNCS20,ALEXANDER2020132520}, the accuracy of these emulators depends on the kernel and the problem of selecting a good kernel has received less attention. 

We investigate Kernel Flows \cite{Owhadi19} (KF) as a generic tool for selecting the kernel used to learn chaotic dynamical systems. The KF strategy is to induce  an ordering (quantifying the quality of a kernel) in a space of kernels and use gradient descent to identify a good kernel. KF is an efficient method of learning kernels with predictive capabilities using random projections that guarantees good performance while reducing computational cost. KF is also  a variant of cross-validation (see discussion in \cite{kfscvs}) in the sense that it operates under the  premise that a kernel must be good if the number of points  used to interpolate the data can be halved without significant loss in accuracy, i.e., the method presented in  \cite{Owhadi19} uses the regression relative error between two interpolants (measured in the RKHS norm of the kernel) as the quantity to minimize. 

 In this paper, we use this metric along two new ones to learn the parameters of the kernel. The first one 
 is the difference between two estimations of the maximal Lyapunov exponent (the second estimator using a random half of the data points of the first).
The second metric is the Maximum Mean Discrepancy (MMD) \cite{JMLR:v13:gretton12a}  computed from two different samples of a time series or between a sample and a subsample of half length.
Our paper is numerical in nature and we refer to \cite{kfscvs} for a rigorous analysis of KF (and comparisons with Empirical Bayes for learning PDEs) and to \cite{yoo2020deep} for its applications to  training neural networks.

The main contributions of this paper are as follows.
\begin{itemize}
    \item We show that combining KF with the kriging of the vector field significantly improves the accuracy of (1) the prediction of chaotic time series (2) the reconstruction of attractors (3) the reconstruction of the dynamics from lower dimensional projections of the state space.
    \item We show that Kernel Mode Decomposition can recover  time delays in the reconstruction of the dynamics.
    \item We introduce  Lyapunov exponents and MMD as two new cross validation metrics for kriging vector fields.
    
\end{itemize}
The remainder of the manuscript is structured as follows.  We describe the problem in Section 2 and propose three cross-validation metrics to learn the parameters of the kernel used for approximating the vector field of the dynamical system. In section 3, we investigate the performance of these methods for the Bernoulli map, the logistic map, the H\'enon map and the Lorenz system.  In the appendix, we recall  optimal recovery theoretical foundations of KF.








\section{The problem and its proposed cross-validation solutions}
Let $x_1,\ldots,x_k,\ldots$ be a time series in $\R^d$. Our goal is to forecast $x_{n+1}$ given the observation of $x_1,\ldots,x_n$. 
We  work under the assumption that this time series can be approximated by a solution of a dynamical system of the form 
\begin{equation}\label{eqjhdbdjehbd}
z_{k+1}=f^\dagger(z_k,\ldots,z_{k-\tau^\dagger+1}),
\end{equation}
where $\tau^\dagger \in \N^*$ and $f^\dagger$ may be unknown. 
Given $\tau \in \N^*$, the approximation of the dynamical can then be recast as that of interpolating $f^\dagger$ from  pointwise measurements
\begin{equation}\label{eqn:fdagger}
f^\dagger(X_k)=Y_k\text{ for }k=1,\ldots,N
\end{equation}
with $X_k:=(x_{k+\tau-1},\ldots,x_k)$, $Y_k:=x_{k+\tau}$ and  $N=n-\tau$.
Given a reproducing kernel Hilbert space\footnote{A brief overview of RKHSs is given in the appendix.} of candidates $\Hc$ for $f^\dagger$, and using the relative error in the RKHS norm $\|\cdot\|_\Hc$ as a loss, the regression of the data $(X_k,Y_k)$ with the kernel $K$ associated with $\Hc$  provides a
minimax optimal approximation \cite{owhadi_scovel_2019}  of  $f^\dagger$ in $ \Hc$.  This interpolant (in the absence of measurement noise) is  
\begin{equation}\label{mean_gp}
f(x)=K(x,X) (K(X,X))^{-1} Y
\end{equation}
where  $X=(X_{1},\ldots, X_{N})$, $Y=(Y_{1},\ldots, Y_{N})$,  $k(X,X)$ for the $N\times N $ matrix with entries $k(X_i,X_i)$, and $k(x,X)$ is the $N$ vector with entries $k(x,X_i)$. This interpolation has also a natural interpretation in the setting of Gaussian process (GP) regression: (1) \eqref{mean_gp} is the  conditional mean of the centered GP $\xi\sim \mathcal{N}(0,K)$ with covariance function $K$ conditioned on $\xi(X_k)=Y_k$, and (2) the interpolation error between $f^\dagger$ and $f$ is bounded by the conditional standard deviation of the GP $\xi$, i.e. \begin{equation}\label{eqkejbdkddbs}
|f^\dagger(x)-f(x)|\leq \sigma(x) \|f^\dagger\|_\Hc\end{equation}
with 
\begin{equation}\label{variance_gp}
\sigma^2(x)=K(x,x)-K(x,X) (K(X,X))^{-1} K(x,X)^T\,.
\end{equation}

Evidently the accuracy of the proposed approach depends on the  kernel $K$ and one of our goals is to also learn that kernel from the data $(X_k,Y_k)$ with Kernel Flows (KF) \cite{Owhadi19}.

 Given a family of kernels $K_\theta(x,x')$ parameterized by $\theta$, the KF algorithm can then be described as follows \cite{Owhadi19, yoo2020deep}:
 \begin{enumerate}
     \item Select random subvectors $X^b$ and $Y^b$ of $X$ and $Y$ (through uniform sampling without replacement in the index set $\{1,\ldots,N\}$)
     \item Select random subvectors $X^c$ and $Y^c$ of $X^b$ and $Y^b$ (by selecting, at random, uniformly and without replacement, half of the indices defining $X^b$)
     \item  Let\footnote{ $\rho:=\|u^b-u^c\|^2_{K_\theta}/\|u^b\|^2_{K_\theta}$, with $u^b(x)=K_\theta(x,X^b) K_\theta(X^b,X^b)^{-1} Y^b$ and $u^c(x)=K_\theta(x,X^c) K_\theta(X^c,X^c)^{-1} Y^c$, and $\rho$  admits  the representation \eqref{eqjehdhebdhdhj} enabling its computation}
 \begin{equation}\label{eqjehdhebdhdhj}
 \rho(\theta,X^b,Y^b,X^c,Y^c):=1-\frac{Y^{c,T} K_\theta(X^c,X^c)^{-1} Y_c}{Y^{f,T} K_\theta(X^b,X^b)^{-1} Y^b}\,,
 \end{equation}
  be the squared relative error (in the RKHS norm $\|\cdot\|_{K_\theta}$ defined by $K_\theta$)  between
 the interpolants $u^b$ and $u^c$ obtained from the two nested subsets of the dataset and the kernel $K_\theta$
    \item Evolve $\theta$ in the gradient descent direction of $\rho$, i.e. $\theta \leftarrow \theta - \delta \nabla_\theta \rho$
    \item Repeat.
 \end{enumerate}

We also  consider different metrics in step 3 of the algorithm described above. The first new metric is by considering, in the case of chaotic systems, that a kernel is good  if the estimate of the Lyapunov exponent obtained from the kernel approximation of the dynamics does not change 
if half of the data is used. So we will minimize \[\label{rho_l}\rho_{L}=|\lambda_{\mbox{max},N}- \lambda_{\mbox{max},N/2}|, \]
instead of \eqref{eqjehdhebdhdhj} with   $\lambda_{\mbox{max},N}$ is the estimate of the maximal Lyapunov exponent from the kernel approximation of the dynamics with $N$ sample points and  $\lambda_{\mbox{max},N/2}$ is the estimate of the maximal Lyapunov exponent from the kernel approximation of the dynamics with $N/2$ sample points. We use the algorithm of  Eckmann et al. \cite{eckmann} to estimate the Lyapunov exponents from data by considering the kernel approximation of the dynamics. We use the Python implementation in \cite{nolds} to estimate the Lyapunov exponents from data.

The second new metric is based on the Maximum Mean Discrepancy (MMD) \cite{JMLR:v13:gretton12a} that is
  a distance on the space of probability measures with a representer theorem for empirical distributions which we recall in the appendix. Our strategy  for learning the kernel $K$ will then simply  be to minimize the MMD 
\begin{equation}
  \label{rho_mmd}
  \rho_{\tiny \mbox{MMD}}=\mbox{MMD}(S_1,S_2)  
\end{equation} 
between two different samples\footnote{One could also consider the MMD
between a sample $S_1$ of size $m$ and a subsample of $S_1$ of size $m/2$.}, $S_1=x_{\sigma_1},\cdots, x_{\sigma_m}$ and $S_2=x_{\mu_1},\cdots, x_{\mu_m}$,  of the time series.

\section{Numerical experiments}
We  now numerically  investigate the efficacy of the cross-validation approaches described in the previous section in learning chaotic dynamical systems.

\subsection{Bernoulli map}
We  first use the Bernoulli map 
\begin{equation}\label{eqkjhdkjehdkjdhgb}
x(k+1)=2x(k) \mbox{ mod } 1\,,
\end{equation}
which is a  prototypical chaotic dynamical system \cite{sebastian}.  
We initialize \eqref{eqkjhdkjehdkjdhgb} from an
  (irrational) initial condition $x(0)= \pi/3$ and use $200$ points to train the kernel and for interpolation. 
We use a parameterized family of kernels of the form    
\begin{equation}
k(x,y)= \alpha_0\, \mbox{max}\{0,1-\frac{||x-y||_2^2|}{\sigma_0}\}+\alpha_1\, e^{\frac{||x-y||_2^2}{\sigma_1^2}}\,
\end{equation}
We set the initial kernel to be the Gaussian kernel and initialize  the parameters with $(\alpha_0,\sigma_0,\alpha_1,\sigma_1)=(0,1,1,1)$. 
The parameters of the kernel after training with $\rho$ and $\rho_{\mbox{MMD}}$ and the Root Mean Square Errors\footnote{The Root Mean Square Error (RMSE) is a standard way to measure the error of a model in predicting quantitative data. Formally it is defined as $\mbox{RMSE}=\sqrt{\frac{\sum_{i=1}^n(\hat{y}_i-y_i)^2}{n}}$ with $\hat{y}_1,\cdots,\hat{y}_n$ are predicted values, $y_1,\cdots,y_n$ are observed values and $n$ is the number of observations.} (RMSEs) with 5,000 points  are summarized in the following table with $R_1$ being the RMSE for $x(0)=\pi/10$ and $R_2$ the RMSE for $x(0)=0.1$.
\begin{center}
\newcolumntype{g}{>{\columncolor{Gray}}c}
\begin{tabular}{ |g |c| c |c| c|} \hline 
      &    $  [\alpha_0,\sigma_0,\alpha_1,\sigma_1]$  &   No. of iterations  &  $R_1$ &  $R_2$  \\ \hline
   $\rho$ & $[1.31,1.01, 0.99,0.99]$  &  100  & $0.019$ & 0.015  \\ \hline
  $\rho_{\mbox{\tiny MMD}}$ &  $[0.830, 2.780, 0.562,2.926]$  &  1000  &  0.027  & 0.011  \\ \hline 
\mbox{No learning} & $[0,1,1,1]$ &   0 &  0.182   & 0.118 \\ \hline 
\end{tabular}
\end{center}

\begin{figure}[t]
\centering
\subfigure[Time series generated by the true dynamics (red) and the approximation (blue) with the learned kernel (left)   and the initial kernel   (right), for an irrational initial condition $\pi/10$.]
{\includegraphics[width=.4\textwidth]{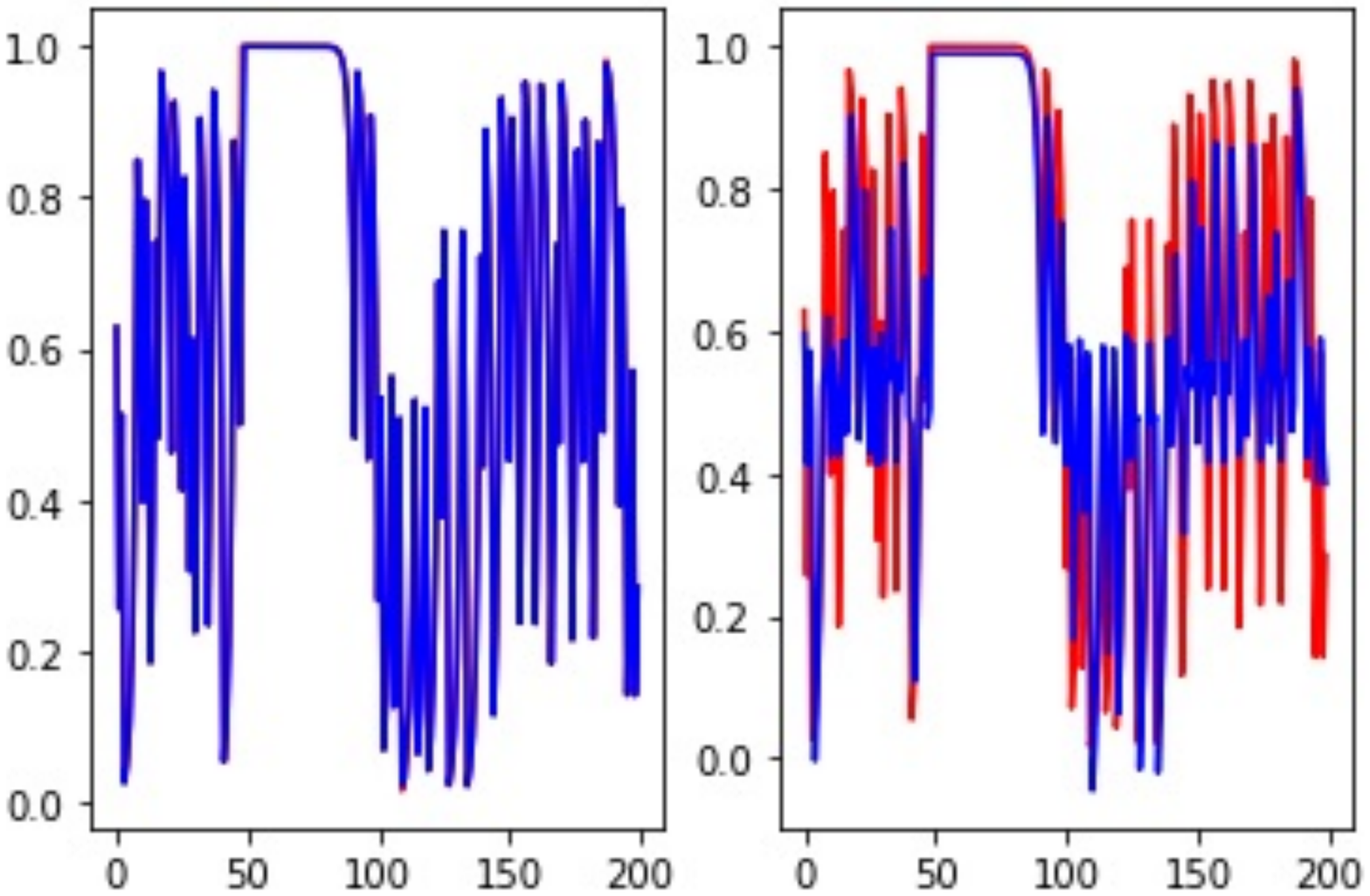}}
\hspace{0.2cm}
\subfigure[Time series generated by the true dynamics (red), the approximation with the learned kernel (blue), the kernel approximation without learning the kernel (green), for a rational initial condition $0.1$]{
\includegraphics[width=.4\textwidth]{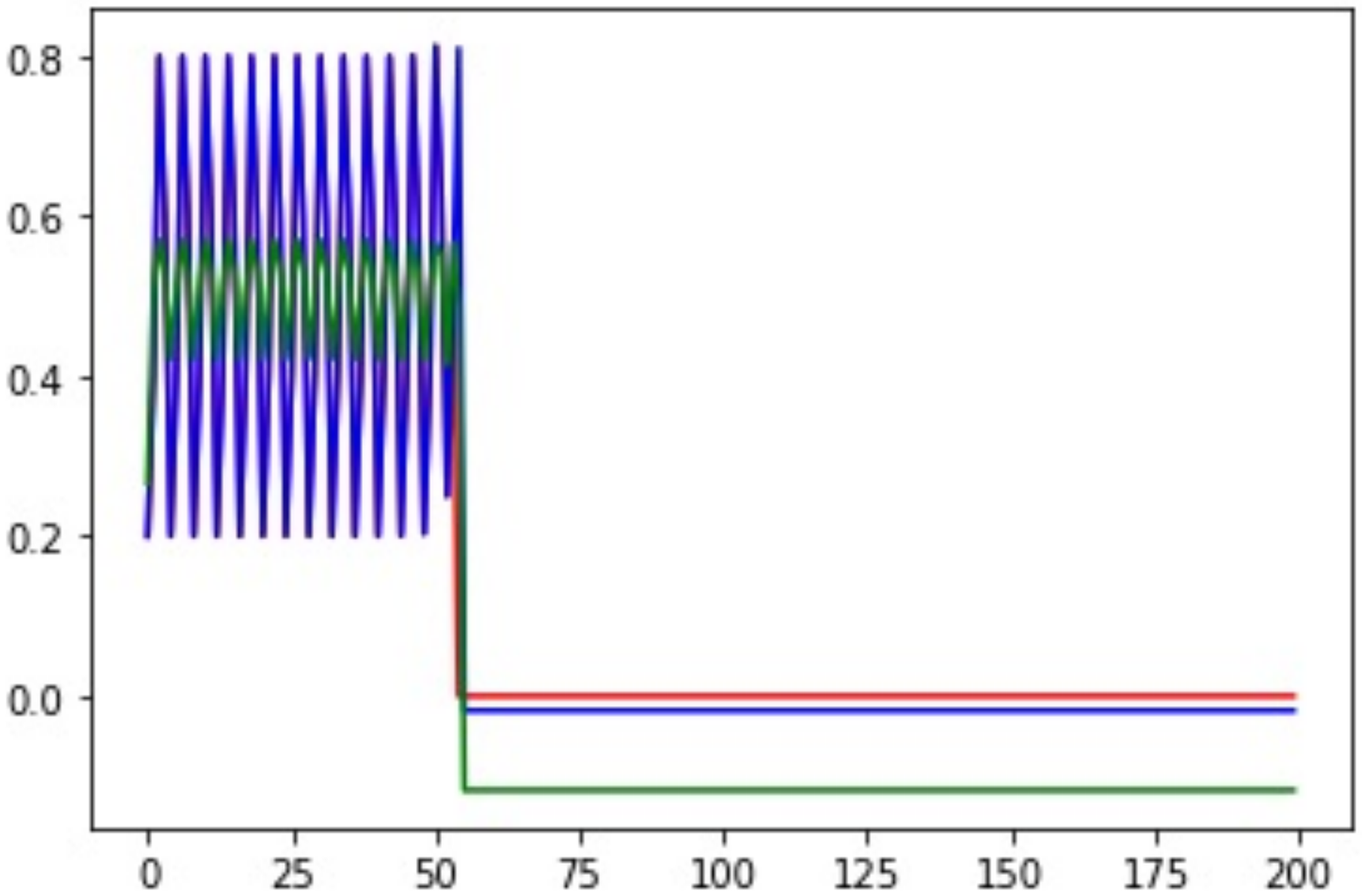}
}
\caption{Time series generated by the true dynamics, approximation using the learned kernel and the kernel without learning for different initial conditions}
\label{bernoulli}
\end{figure}

 \noindent Figure \ref{bernoulli} shows  results for an irrational initial condition $x(0)= \pi/10$ and 5000 points and a rational initial condition $x(0)= 0.1$. 

 We also consider a parameterized family of kernels of the form    
\begin{equation}
k(x,y)= 
\alpha_0\, \mbox{max}\{0,1-\frac{||x-y||_2^2|}{\sigma_0}\}+
\alpha_1\, e^{\frac{||x-y||_2^2}{\sigma_1^2}}+\alpha_2 e^{-\frac{||x-y||_2}{\sigma_2^2}}
+\alpha_3
 e^{- \sigma_3  \sin^2(\sigma_4 \pi ||x-y||_2)}e^{- \frac{||x-y||_2^2}{\sigma_5^2}}
+\alpha_4  ||x-y||_2^2
\end{equation}
Results are summarized in the following table
\begin{center}
\newcolumntype{g}{>{\columncolor{Gray}}c}
\begin{tabular}{ |g |c| c |c| c|} \hline 
      &    $  [\alpha_0,\sigma_0,\alpha_1,\sigma_1,\alpha_2,\sigma_2,\alpha_3,\sigma_3,\sigma_4,\sigma_5, \alpha_5]$  &   No. of it.  &  $R_1$ &  $R_2$  \\ \hline
   $\rho$ & $[23.98, 1.13,1.13, 0.83, 32.73,0.72, 32.09,0.29,4.47,0.20,
 0.10]$  &  500  & $0.016$ & 0.014  \\  \hline 
\mbox{No learning} & $[0,1,1,1,0,1,0,1,1,1,0]$ &   0 &  0.182   & 0.118 \\ \hline 
\end{tabular}
\end{center}

\subsection{Example 2 (Logistic map):} Consider the logistic map $x(k+1)=4 x(k)(1-x(k))$. To approximate this map, we use an  initial condition $x(0)= 0.1$ and use 200 points to train the kernel and for interpolation. 
We use a kernel of the form $$k(x,y)=\alpha_0 e^{-\sigma_1\sin^2(\pi \sigma_2 ||x-y||_2^2)} e^{-||x-y||_2^2/\sigma_3^2}$$ and initialize with the set of parameters $(\alpha_0,\sigma_1,\sigma_2,\sigma_3)=(1,1,1,1)$.
Let  $R_1$ be the RMSE  for an initial condition $x(0)=0.4$, $R_2$ for $x(0)=0.97$ with 5000 points.


\begin{center}
\newcolumntype{g}{>{\columncolor{Gray}}c}
\begin{tabular}{ |g |c| c |c| c|} \hline 
      &    $  [\alpha_0,\sigma_1,\sigma_2,\sigma_3]$  &   No. of it.  &  $R_1$ &  $R_2$  \\ \hline
   $\rho$ & $[0.95, 0.98,1.20,0.62]$  &  100  &  0.0004 & 0.002 \\ \hline
   $\rho_L$ & $[0.6, 1.8,2.3,1.4]$  &  1000  & 0.001 & 0.001  \\  \hline 
\mbox{No learning} & $[1,1,1,1]$ &   0 &  0.004   & 0.0004 \\ \hline 
\end{tabular}
\end{center}

\begin{figure}[t]
\centering
\subfigure[Time series generated by the true dynamics (red) and the approximation with the learned kernel using $\rho$ (blue), the approximation with the learned kernel using $\rho_L$ (green), approximation without learning (yellow)]{
\includegraphics[width=.45\textwidth]{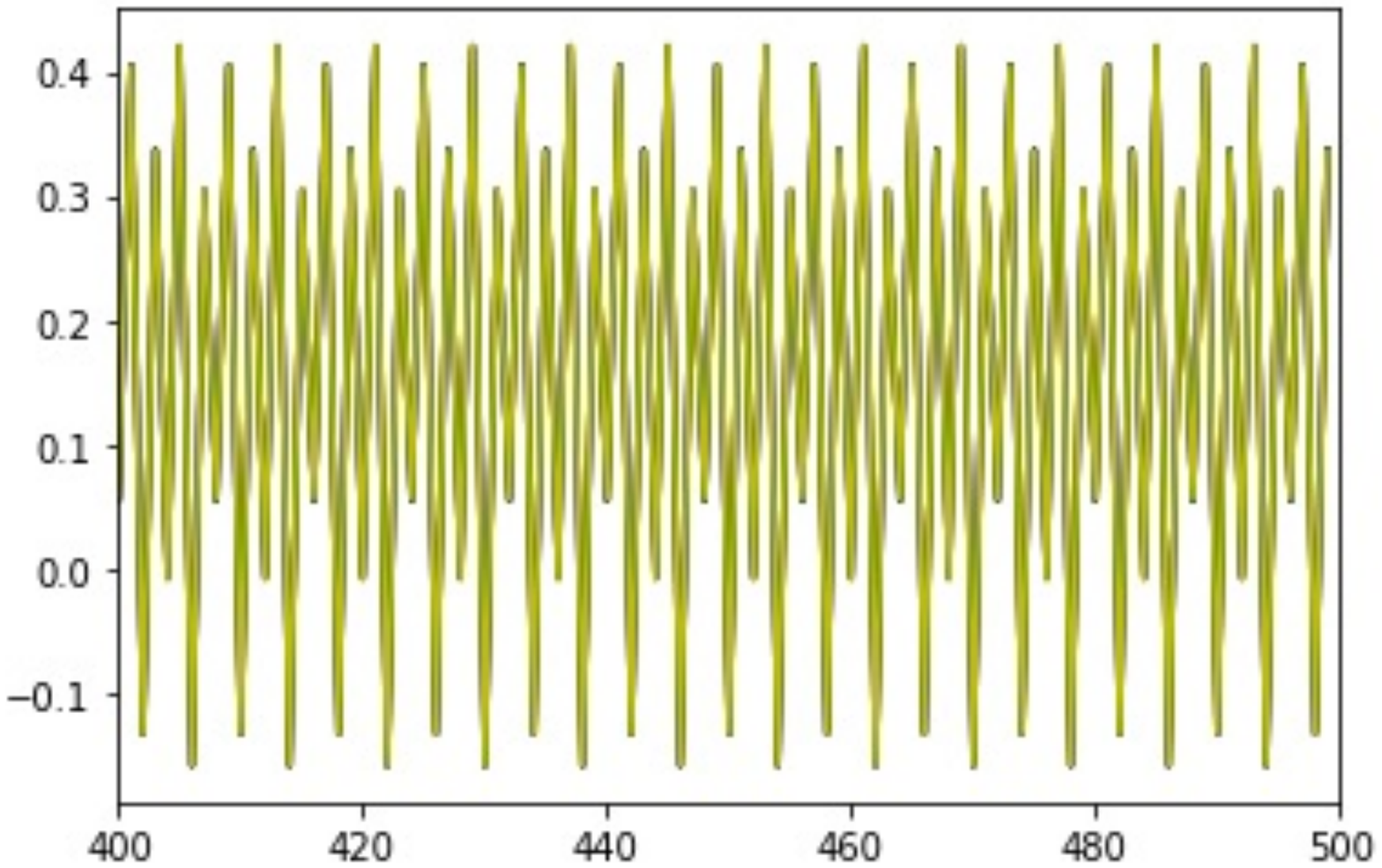}
}
\hspace{0.4cm}
\subfigure[Difference between the true and the approximated dynamics with the learned kernel using $\rho$  (top), with the learned kernel using $\rho_L$ (middle), with the initial kernel (bottom), for an initial condition $x_0=0.3$]{
\includegraphics[width=.45\textwidth]{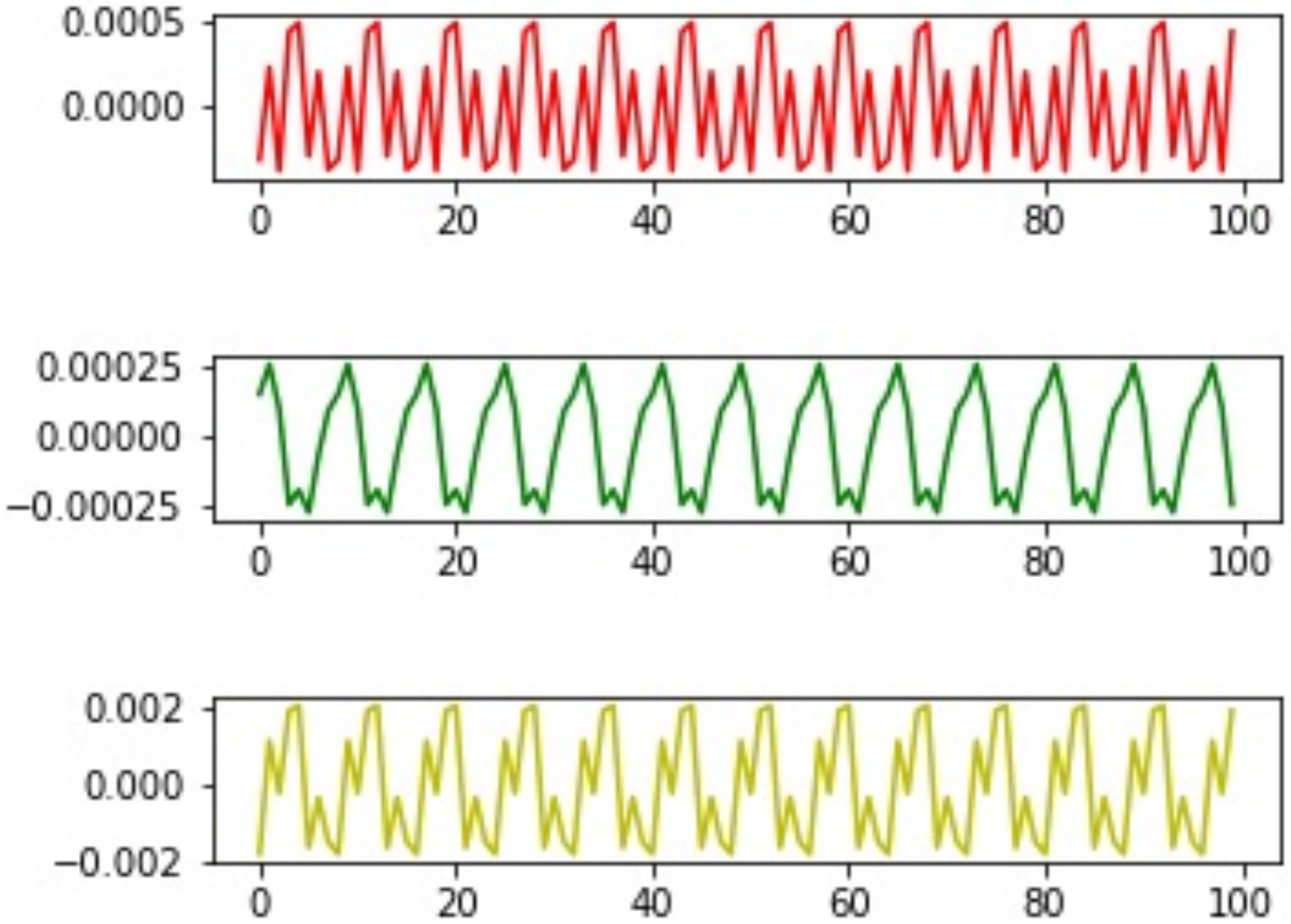}
}
\caption{Prediction results for the logistic map}
\label{logistic}
\end{figure}

\begin{figure}[h]
    \centering
    \includegraphics[width=0.5\textwidth]{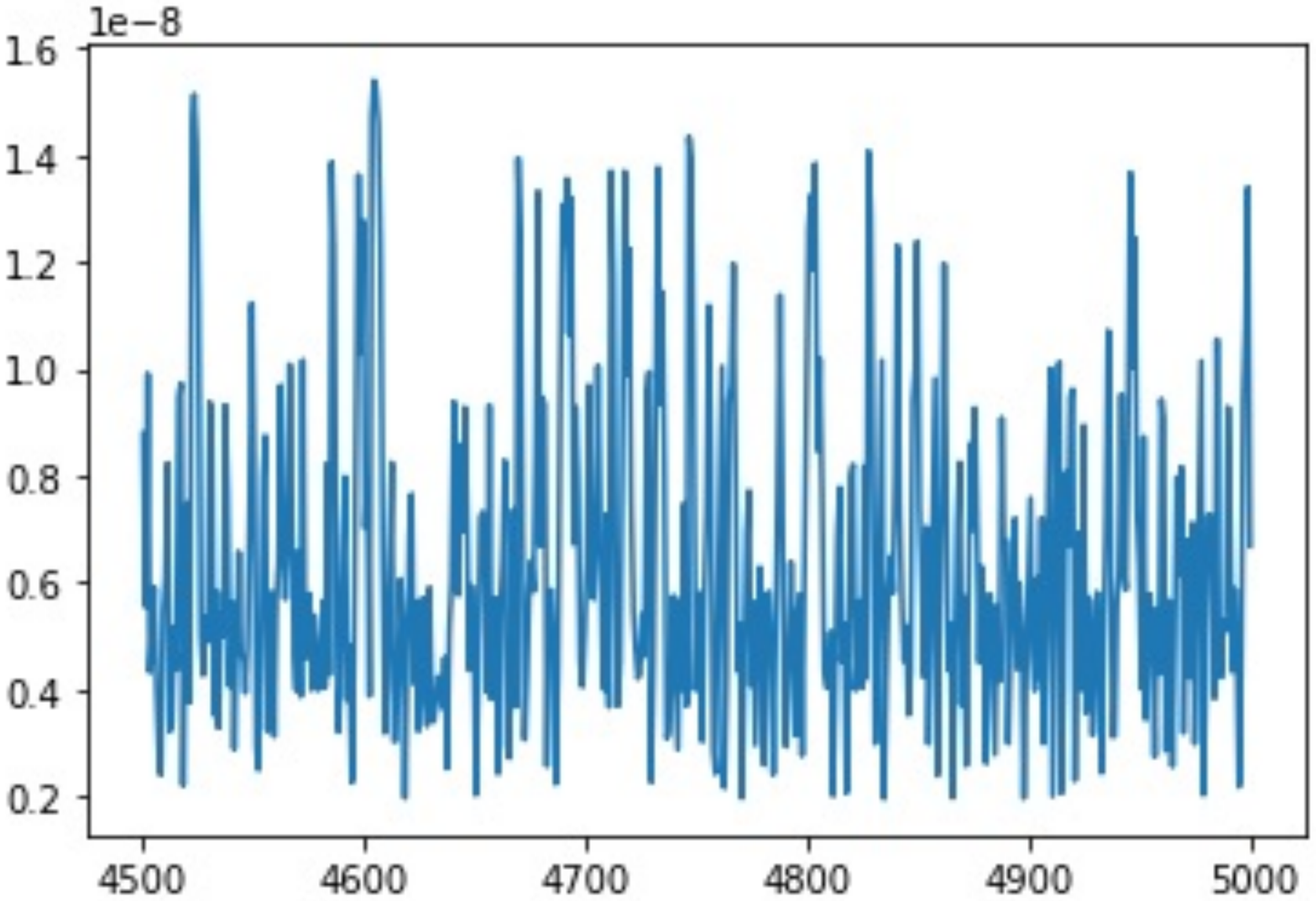}
\caption{Uncertainty $\Delta(f(x))$ in formula (\ref{error_estimate}) for an initial condition $x_0=\pi/4$}
 \label{variance_logistic}
\end{figure}

\noindent Figure \ref{logistic}.a shows the results for an initial condition $x(0)= 0.3$ and 5000 points. Figure \ref{logistic}.b shows the prediction errors for the case of an approximation with a learned kernel using $\rho$, $\rho_L$ and a kernel without learning.
  Figure \ref{variance_logistic} shows the plot of error interval for $f^\dagger(x)$ given by $\Delta(f(x))$ in (\ref{delta}). 
 
 We also consider a parameterized family of kernels of the form    
\begin{equation}
k(x,y)= 
\alpha_0^2\, \mbox{max}\{0,1-\frac{||x-y||_2^2|}{\sigma_0}\}+
\alpha_1^2\, e^{\frac{||x-y||_2^2}{\sigma_1^2}}+\alpha_2^2 e^{-\frac{||x-y||_2}{\sigma_2^2}}
+\alpha_3^2
 e^{- \sigma_3  \sin^2(\sigma_4 \pi ||x-y||_2^2)}e^{- \frac{||x-y||_2^2}{\sigma_5^2}}
+\alpha_4^2  ||x-y||_2^2
\end{equation}
We initialize with a gaussian kernel. The results are summarized in the following table where $R_1$ corresponds to the RMSE with $x(0)=0.4$ and $R_2$ corresponds to the RMSE with $x(0)=0.97$.
\begin{center}
\newcolumntype{g}{>{\columncolor{Gray}}c}
\begin{tabular}{ |g |c| c |c| c|} \hline 
      &    \scriptsize $   [\alpha_0,\sigma_0,\alpha_1,\sigma_1,\alpha_2,\sigma_2,\alpha_3,\sigma_3,\sigma_4,\sigma_5,\alpha_4]$  &   No. of it.  &  $R_1$ &  $R_2$  \\ \hline
  $\rho$ & \scriptsize $[ 0.15,
 0.96,
 0.99,
  1.02,
  0.08,
  0.98,
  -3.96\, 10^{-05},
  0.99,
  0.99,
  0.99,
  0.98]$  &  500  &  0.0003 & 0.0004  \\  \hline 
\mbox{No learning} & $[0,1,1,1,0,1,0,1,1,1,0]$ &   0 &  0.004   & 0.004 \\ \hline 
\end{tabular}
\end{center}

\subsection{Example 3 (H\'enon map)}
Consider the H\'enon map
\begin{eqnarray*}
x(k+1) &=& 1-ax(k)^2+y(k) \\
y(k+1) &= & b x(k)
\end{eqnarray*}
with $a = 1.4$ and $b = 0.3$. To learn this map, we generate 100 points with initial conditions $(x(0), y(0))=(0.9, -0.9)$ to learn two kernels $$k_i(x,y)=\alpha_{i}+(\beta_{i}+||x-y||^{\kappa_i}_2)^{\sigma_i}+\delta_i e^{-||x-y||_2^2/{\mu_i^2}}$$ ($i=1,2$) corresponding to the two maps $\left[\begin{array}{c}x(k)\\y(k) \end{array} \right] \mapsto x(k+1)$ and $\left[\begin{array}{c}x(k)\\y(k) \end{array} \right] \mapsto y(k+1)$. We initialize with a gaussian kernel and after 1000 iterations, we get\footnote{We notice that the algorithm converges to non-integer powers. Terms of the form $||x-y||_2^{\alpha}$ can be represented as $e^{\alpha \log||x-y||_2}$ which could be a reproducing kernel. }

\begin{center}
\newcolumntype{g}{>{\columncolor{Gray}}c}
\begin{tabular}{ |g |c| c |c| } \hline 
      &    $\left[\begin{array}{ccccccc}\alpha_{1} & \beta_{1}& \kappa_1 & \sigma_1 & \delta_1 & \mu_1\\
\alpha_{2} & \beta_{2} & \kappa_2 & \sigma_2 &  \delta_2& \mu_2     \end{array}\right]$  &   No. of it.  &  $R_1$   \\ \hline
   $\rho$ & $\left[\begin{array}{ccccccc}0.99 & 1.12 &  0.74 & 2.21 & 0.98 & 0.89\\
   1.00 & 1.01 &  3.35 & 0.008 & 0.95 & 1.35
    \end{array}\right]$   &  1000  &  $\left[\begin{array}{c} 0.04 \\
0.01 \end{array}\right]$   \\  \hline 
\mbox{No learning} & $\left[\begin{array}{ccccccc}0.0 & 0.0 & 0.0 & 0.0 & 1.0 & 1.0   \\
0.0 & 0.0 & 0.0 & 0.0 & 1.0 & 1.0     \end{array}\right]$ &   0 &  $\left[\begin{array}{c} 0.07 \\
0.01 \end{array}\right]$    \\ \hline 
\end{tabular}
\end{center}

\begin{figure}[t]
\centering
\subfigure[True (blue) and approximated dynamics with the learned kernel (red) ($x-$ component on the left, $y-$ component on the right)]{
\includegraphics[width=.46\textwidth]{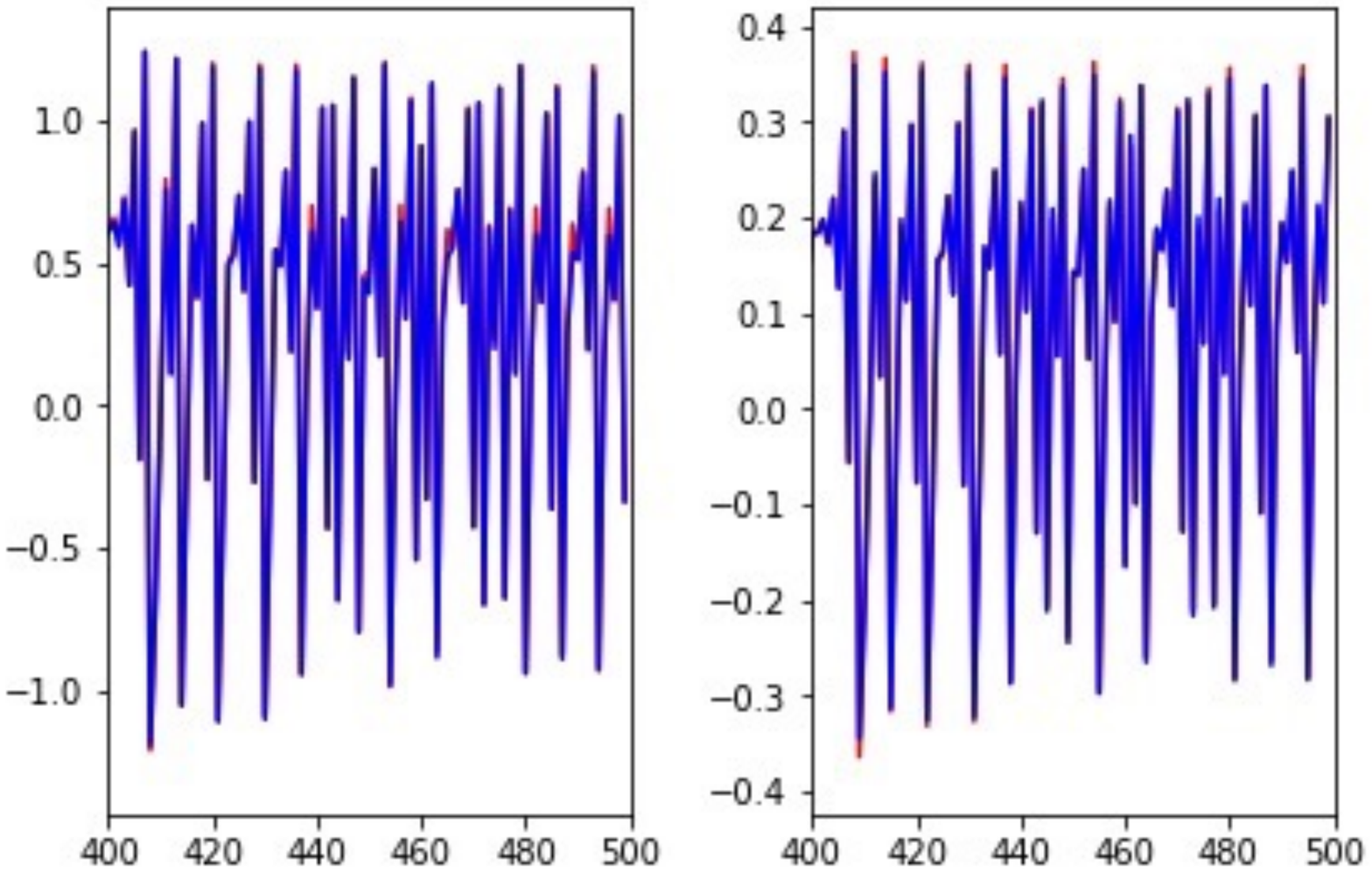}
}
\hspace{0.4cm}
\subfigure[Difference between the true and the approximated dynamics with the learned kernel (blue),  with the initial kernel (red)  ($x-$ component on the left, $y-$ component on the right)]{
\includegraphics[width=.46\textwidth]{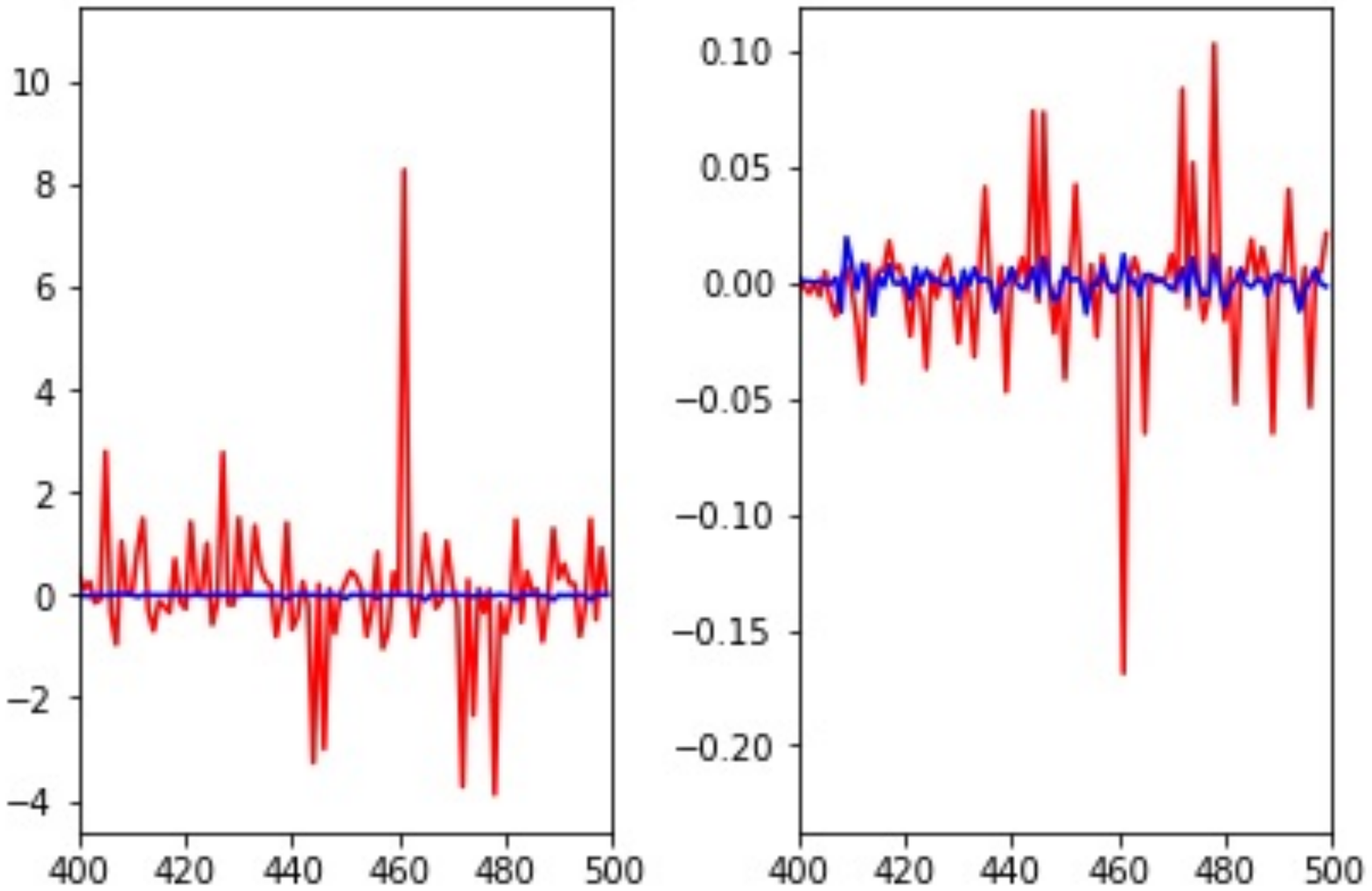}
}
\caption{Prediction results for the H\'enon map}
\label{henon1}
\end{figure}

\noindent We generate a time series for the initial conditions $(x(0),y(0))=(-0.1,0.1)$ and simulate for 5000 points. Figure \ref{henon1} shows the true and approximated dynamics as well as the difference between the true and approximated dynamics using the learned kernel and without learning the kernel. \\
We also consider a parameterized family of kernels of the form    
\begin{equation}
k(x,y)= 
\alpha_{0,i}^2\, \mbox{max}\{0,1-\frac{||x-y||_2^2|}{\sigma_{0,i}}\}+
\alpha_{1,i}^2\, e^{\frac{||x-y||_2^2}{\sigma_{1,i}^2}}+\alpha_{2,i}^2 e^{-\frac{||x-y||_2}{\sigma_{2,i}^2}}
+\alpha_{3,i}^2
 e^{- \sigma_{3,i}  \sin^2(\sigma_{4,i} \pi ||x-y||_2^2)}e^{- \frac{||x-y||_2^2}{\sigma_{5,i}^2}}
+\alpha_{4,i}^2  ||x-y||_2^2
\end{equation}
We initialize with a gaussian kernel. The results are summarized in the following table where $R_1$ corresponds to the RMSE with $x(0)=0.4$ and $R_2$ corresponds to the RMSE with $x(0)=0.97$  and 5000 points.
\hspace{-1cm}\begin{center}
\newcolumntype{g}{>{\columncolor{Gray}}c}
\begin{tabular}{ |g |c| c |c| } \hline 
      &   $\scriptsize \left[\begin{array}{ccccccccccc}
      \alpha_{0,1}&\sigma_{0,1}&\alpha_{1,1}&\sigma_{1,1}&\alpha_{2,1}&\sigma_{2,1}&\alpha_{3,1} & \sigma_{3,1} & \sigma_{4,1} & \sigma_{5,1} & \alpha_{4,1}\\
      \alpha_{0,2}&\sigma_{0,2}&\alpha_{1,2}&\sigma_{1,2}&\alpha_{2,2}&\sigma_{2,2}&\alpha_{3,2} & \sigma_{3,2} & \sigma_{4,2} & \sigma_{5,2} & \alpha_{4,2}     \end{array}\right]$  &   N  &  $R_1$   \\ \hline
   $\rho$ & $\scriptsize \left[\begin{array}{ccccccccccc}  4.48\, 10^{-08} & 1.00 & 2.25 & 2.41 & 0.0& 1.01& 0.17& 1.07& 1.17 & 1.21 & 0.60 \\
 0.18 & 0.96 & 1.09 & 2.30 & 0.20 & 1.00 & 0.26 & 1.03 & 1.11 & 0.84 & 1.65 \, 10^{-14}     \end{array}\right]$   &  \scriptsize 5000  &  $\scriptsize \left[\begin{array}{c}0.05\\
0.008 \end{array}\right]$   \\  \hline 
\mbox{No learning} & $\scriptsize \left[\begin{array}{ccccccccccc}0.0 & 1.0 & 1.0 & 1.0 & 0.0 & 1.0 & 0.0 & 1.0 & 1.0 & 1.0 & 0.0 \\
0.0 & 1.0 & 1.0 & 1.0 & 0.0 & 1.0 & 0.0 & 1.0 & 1.0 & 1.0 & 0.0   \end{array}\right]$ &   0 &  $\scriptsize \left[\begin{array}{c}0.08\\
0.01 \end{array}\right]$    \\ \hline 
\end{tabular}
\end{center}

\subsubsection{Finding $\tau$}
\begin{figure}[t]
\centering
\subfigure[The RMSE as a function of $\tau$]{
\includegraphics[width=.3\textwidth]{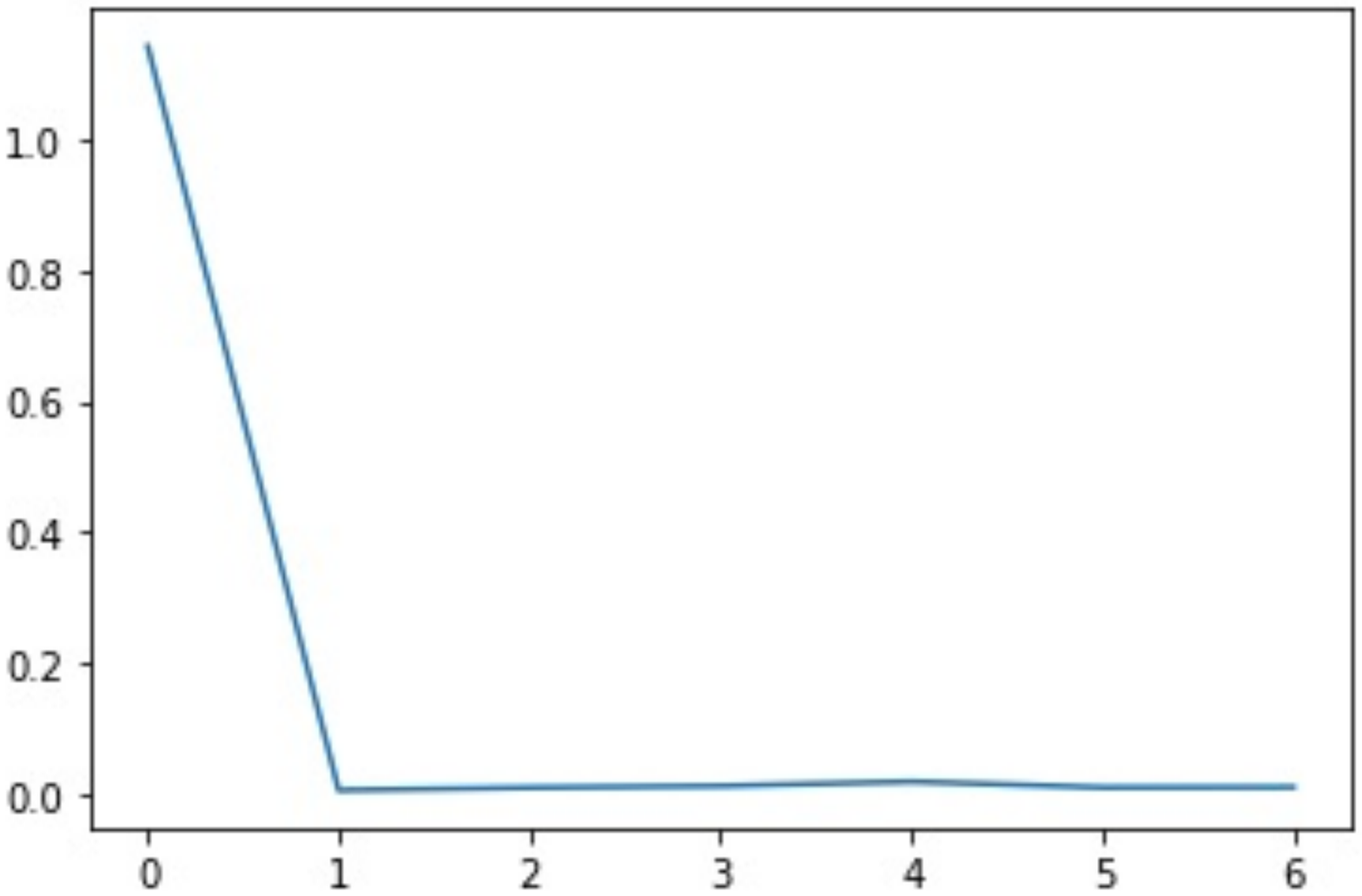}}
 \hspace{0.2cm}
\subfigure[Energy of the time-delays using KMD for the $x-$dynamics]{
\includegraphics[width=.3\textwidth]{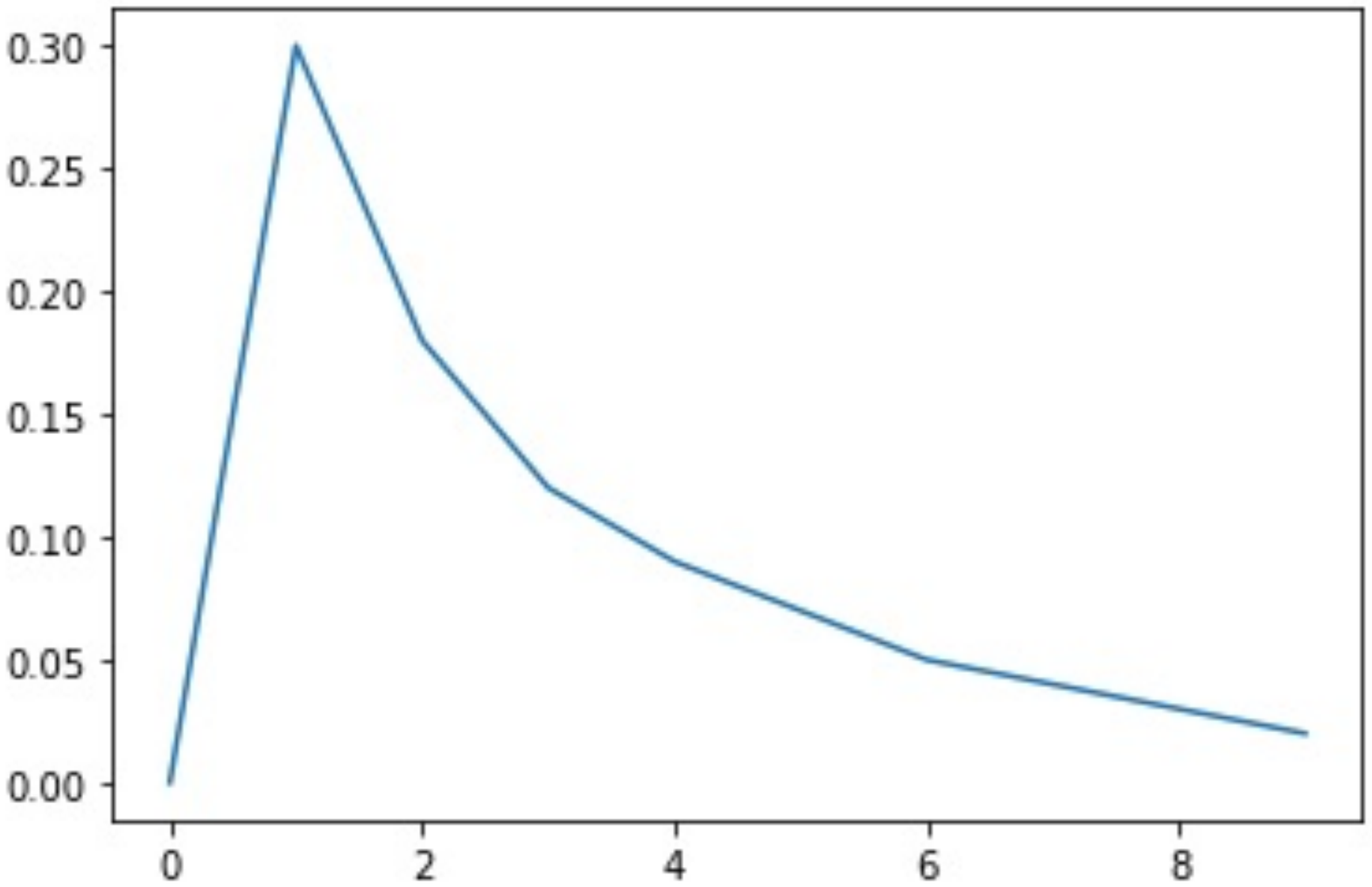}}
\hspace{0.2cm}
\subfigure[Energy of the time-delays using KMD for the $y-$dynamics]{
\includegraphics[width=.3\textwidth]{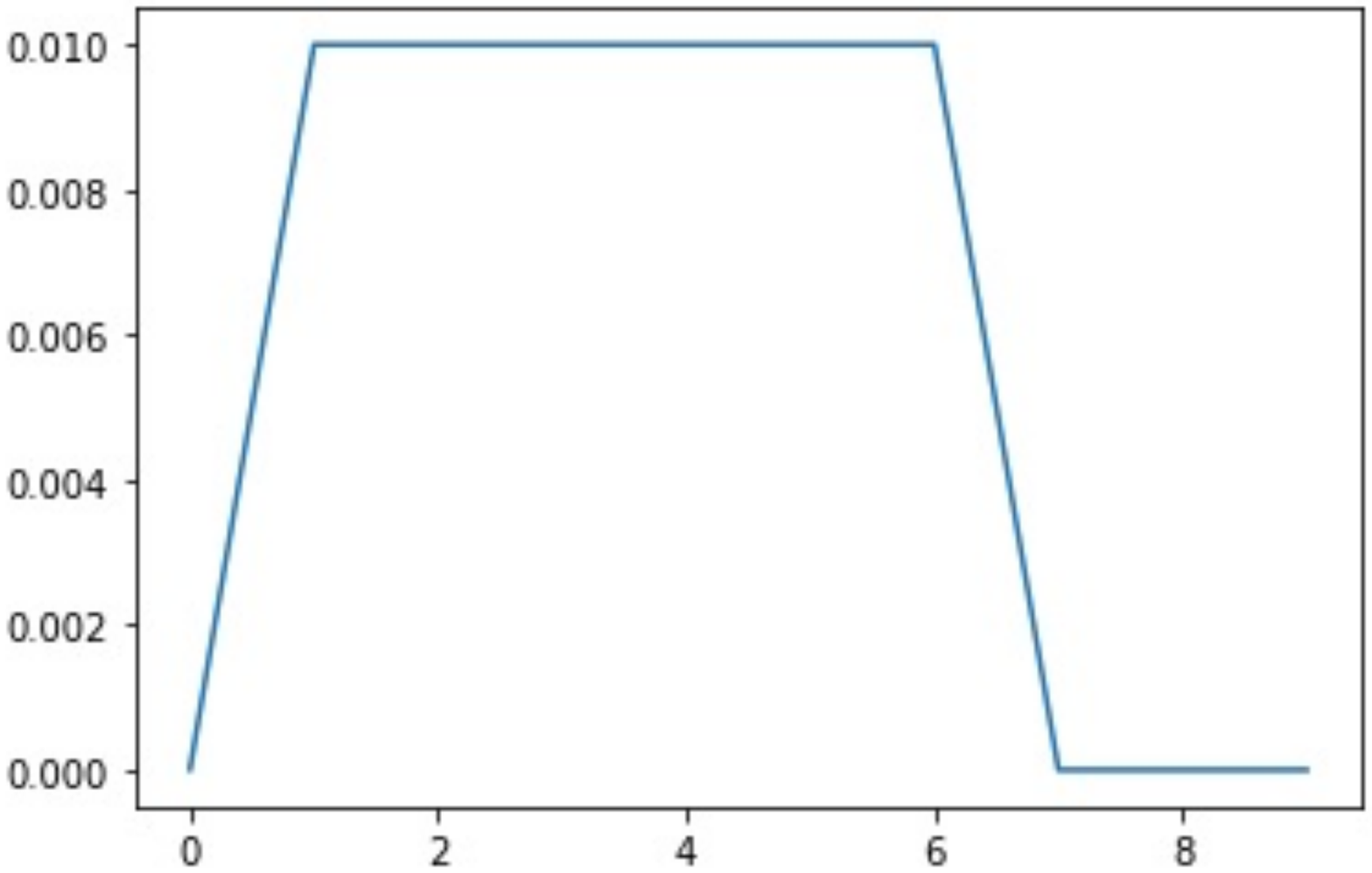}
}
\caption{Energy of the time-delays using  RMSE and KMD}
\label{henon_energies_kmd}
\end{figure}

Now, we consider the scalar dimensional version of the H\'enon map as $x(k+1)=1-ax(k)^2+bx(k-1)$.  We aim at learning the kernel and finding the optimal time delay $\tau$. We start with an initial condition $(x(0), y(0))= (0.8, -0.9)$ and generate 100 points for learning. We use a kernel of the form $$k(x,y)=\alpha_0+(\beta_0+||x-y||_2^{\gamma_0})^{\sigma_0}.$$ We generate 100 points for different values of $\tau$ from 0 to 6. Figure \ref{henon_energies_kmd} shows the root mean square error (RMSE) for prediction with 5000 points and initial condition $(x(0), y(0))= (0.1, -0.1)$. It shows that $\tau=1$ is where the RMSE starts stabilizing and can be viewed as an optimal embedding delay.

Another method for finding the embedding delay is the Kernel Mode Decomposition (KMD) \cite{kmd_owhadi} of the time series. We consider a representation of the time series as
\begin{equation}
    v(t+1)=\sum_{j=0}^{N}\alpha_j K(V_{\tau^{\dagger}}(t),V_{\tau^{\dagger}}(j)),
\end{equation}
with $V_{\tau^{\dagger}}(t)=[v(t)\cdots v(t-\tau^{\dagger})]$.
Following \cite{kmd_owhadi}, we define the model alignment energy ${\cal E}_i$ associated to the time-shift  $\tau=i$, $i=0,\cdots,\tau_{\scriptsize \mbox{max}}$ as
\begin{equation}
    {\cal E}_i=v^T K^{-1} K_i K^{-1} v
\end{equation}
with \begin{equation} K(x,y)=\sum_{i=0}^{\tau_{\scriptsize \mbox{\tiny max}}} K_i(x,y)\end{equation}
and $K_i(x,y)=K({\cal S}_i x,{\cal S}_i y)$ with ${\cal S}_i$ the time-truncation operator that truncates time-series at the $i-$th element: given a time series $Y = \{Y_t: t \in \mathbb{T}\}$, where $\mathbb{T}$ is the index set, ${\cal S}_iY=\{[y(t-i)\cdots y(t)]: t \in \mathbb{T}\}$.

We use the embedding delay $\tau^{\dagger}$ that maximizes ${\cal E}_i$.
We apply this method to $x(k+1)=1-ax(k)^2+bx(k-1)$.
We use $K(x,y)=1+e^{-||x-y||_2^2}$ to compute the energies of the embedding delays and get that ${\cal E}_1$ is the maximal value and we deduce that the optimal embedding delay is $1$ which agrees with the model. 

Considering the H\'enon map in the $y-$variable, we get $y(k+2)=b-\frac{a}{b} \, y^2(k+1)+by(k)$.
We compute the energy ${\cal E}_i$ of the embedding delay $i$, observe that ${\cal E}_1$ is the maximal value and  deduce that the optimal embedding delay is $1$ which agrees with the model.

Figure \ref{henon_energies_kmd} shows the values of the energies of the time-delays for both the $x-$ dynamics and $y-$dynamics.
 

\subsubsection{Using partial information to approximate the dynamics}
\begin{figure}[t]
\centering
\subfigure[$x-$component]{
\includegraphics[width=.35\textwidth]{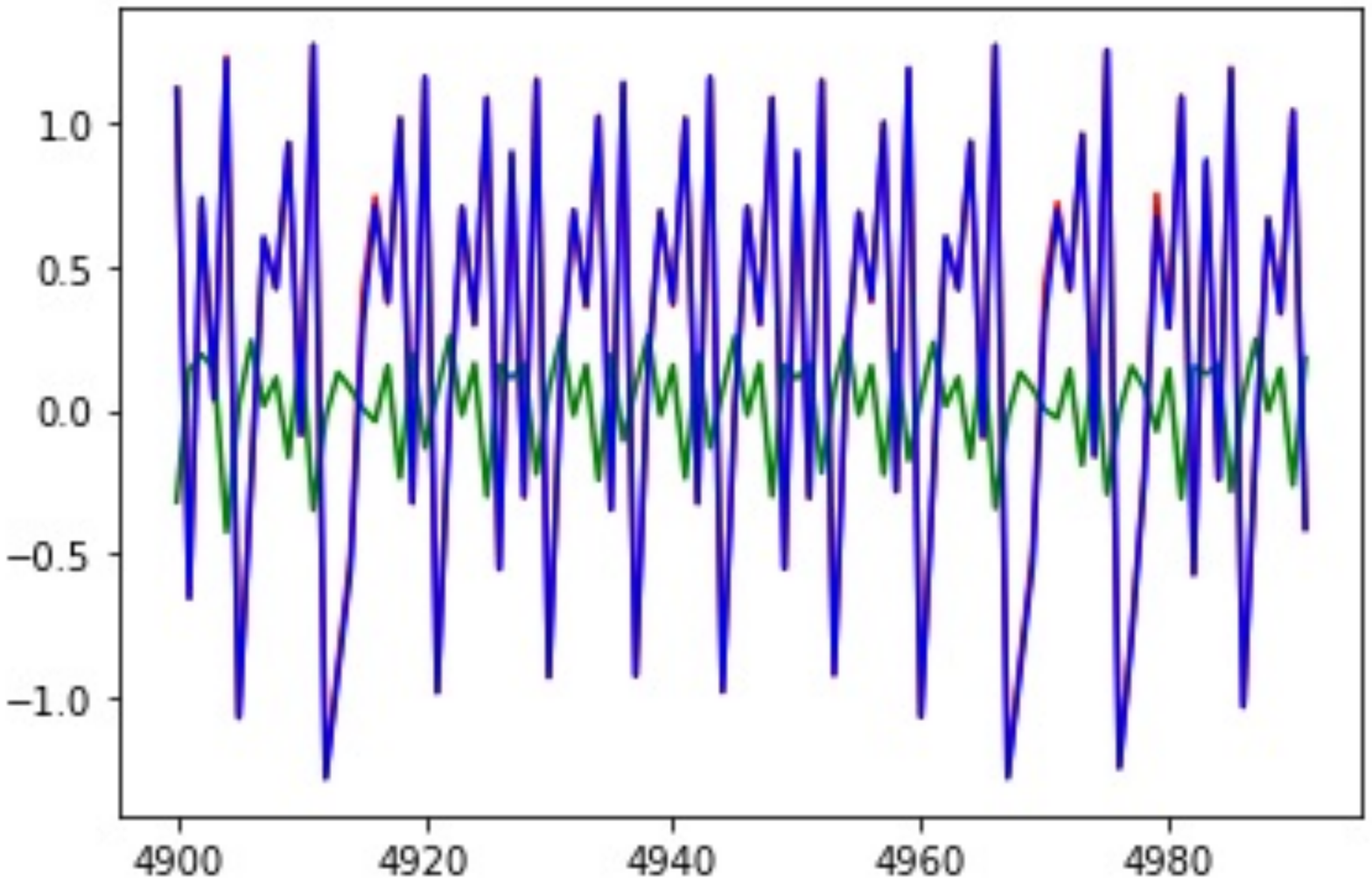}
}
\subfigure[$y-$component]{
\includegraphics[width=.35\textwidth]{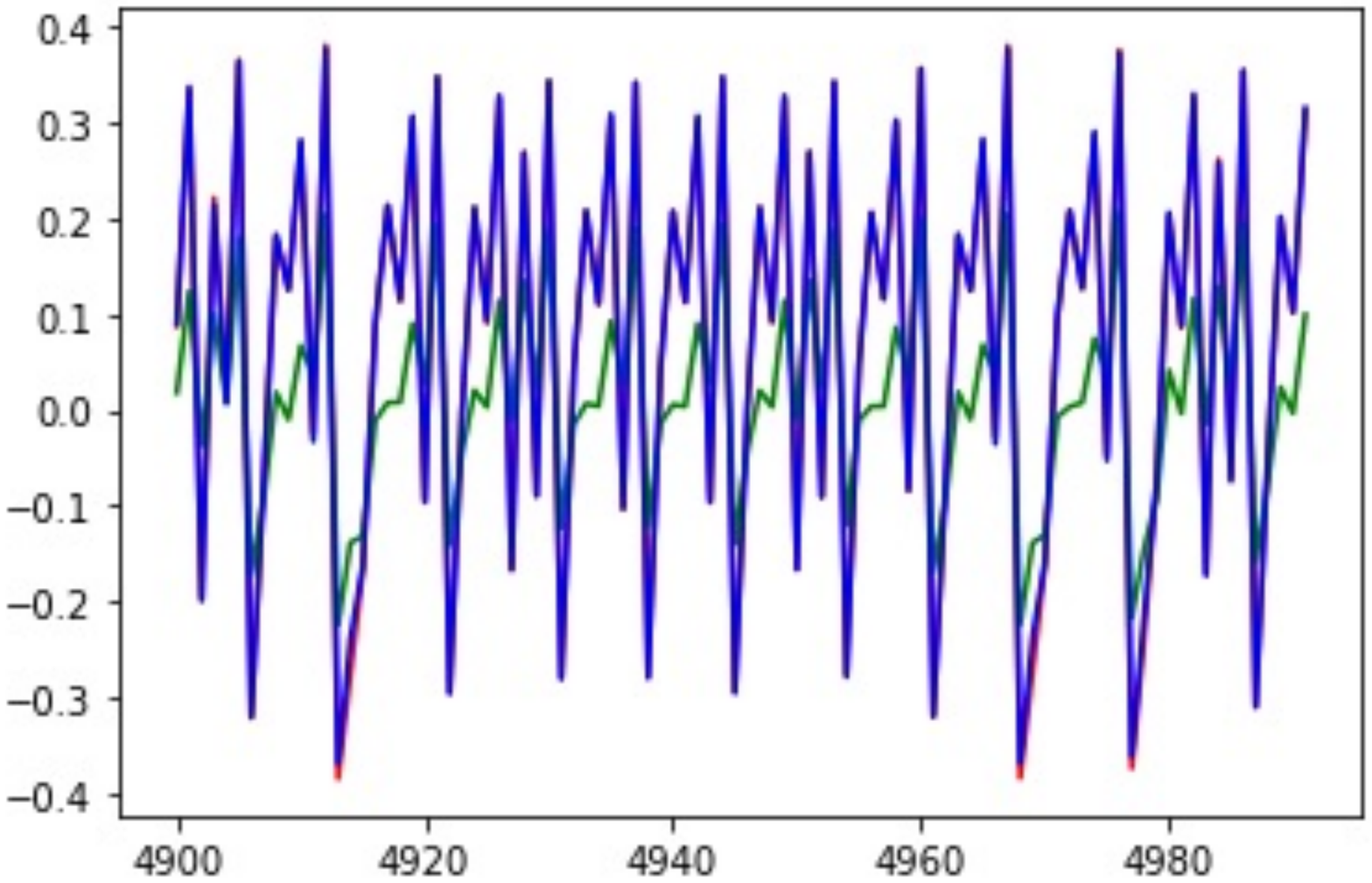}
}

\caption{True dynamics (red), approximated dynamics with the learned kernel (blue), with the kernel without learning (green)}
\label{henon1p}
\end{figure}

In order to learn the dynamics with partial information using measurements from $x$ only, we use the kernel
$$k_i(x,y)= \alpha_{1,i}^2\max(0,1-\frac{||x-y||^2}{\sigma_{1,i}})
+\alpha_{2,i}^2e^{-\frac{||x-y||^2}{\sigma_{2,i}^2}}
+\alpha_{3,i}^2||x-y||^2+\alpha_{4,i}^2e^{-\frac{||x-y||}{\sigma_{4,i}}},
  $$ 
and $\tau=1$, i.e. we learn kernels for the mappings $\left(\begin{array}{c}x(k)\\x(k-1) \end{array}\right) \mapsto x(k+1)$ and $\left(\begin{array}{c}x(k)\\x(k-1) \end{array}\right) \mapsto y(k+1)$. We use 50 points with initial condition $x(0),x(1)=(0.9, -0.9)$ for training and the parameters of the learned kernel are summarized in the following table. Figure \ref{henon1p} shows the results  for initial conditions $(x(0),x(1))=(-0.83,0.57)$ with RMSE   $R_1$. 

\begin{center}
\newcolumntype{g}{>{\columncolor{Gray}}c}
\begin{tabular}{ |g |c| c |c| } \hline 
      &  $\left[\begin{array}{ccccccc}\alpha_{1,1}&\sigma_{1,1}&\alpha_{2,1}&\sigma_{2,1}&\alpha_{3,1}&\sigma_{3,1}&\alpha_{4,1}\\
\alpha_{1,2}&\sigma_{1,2}&\alpha_{2,2}&\sigma_{2,2}&\alpha_{3,2}&\sigma_{3,2}&\alpha_{4,2}      \end{array}\right]$  &   No. of it.  &  $R_1$   \\ \hline
   $\rho$ & $\left[\begin{array}{ccccccc}1.5\,10^{-15} & 1.0 & 7.02 & -2.94 & -6.75 & 4.9\, 10^{-47} & 0.07\\
0.21 & 0.75 & 1.70 & 3.54 & 3.7\, 10^{-27} & 0.13 & 0.91     \end{array}\right]$   &  5000  &  $\left[\begin{array}{c}0.019\\
0.005 \end{array}\right]$   \\  \hline 
\mbox{No learning} & $\left[\begin{array}{ccccccc}0.0 & 1.0 & 1.0 & 1.0 & 0.0 & 1.0 & 1.0 \\
0.0 & 1.0 & 1.0 & 1.0 & 0.0 & 1.0 & 1.0     \end{array}\right]$ &   0 &  $\left[\begin{array}{c}0.87\\
0.14 \end{array}\right]$    \\ \hline 
\end{tabular}
\end{center}

\subsection{Example 4 (The Lorenz system):}

Consider the Lorenz system
\begin{eqnarray}
    \frac{dx}{dt} &=& s (y - x) \\
    \frac{dy}{dt} &=& r x - y - x z \\
    \frac{dz}{dt} &=& x y - bz
\end{eqnarray}    
with $s=10$, $r=28$, $b=10/3$.  We use the  initial condition $(x(0),y(0),z(0))= (0., 1., 1.05)$ and generate 10,000 (training) points with a time step $h=0.01$. 

We randomly pick $N=100$ points out of the original 10,000 points to train the kernel at each iteration (i.e. at each iteration we use 100 randomly selected points to compute the gradient of $\rho$ and move the parameters in the gradient descent direction by one small step) and use the last random selection of $N=100$ points for interpolation (prediction). We use a kernel of the form $$    K_i(x,y)=\alpha_{0,i}+(\alpha_{1,i}+||x-y||_2)^{\beta_i}+\alpha_{2,i} e^{(-||x-y||_2^2/{\sigma_i^2})}
$$ for $i=1,2,3$. The table below summarizes the results for training  using $\rho$ and $\rho_L$ as well as the RMSE for an initial condition $(x(0),y(0),z(0))= (0.5, 1.5, 2.5)$ and 50,000 points

\hspace{-1cm}\begin{center}
\newcolumntype{g}{>{\columncolor{Gray}}c}
\begin{tabular}{ |g |c| c |c| } \hline 
      &    $\scriptsize \left[\begin{array}{ccccccccccc}
      \alpha_{0,1}&\alpha_{1,1}&\beta_1&\alpha_{2,1}&\sigma_{1}\\ \alpha_{0,2}&\alpha_{1,2}&\beta_2&\alpha_{2,2}&\sigma_{2}\\\alpha_{0,3}&\alpha_{1,3}&\beta_3 &\alpha_{2,3}&\sigma_{3}      \end{array}\right]$  &   \mbox{No. of iterations}  &  $R_1$   \\ \hline
   $\rho$ & $\scriptsize \left[\begin{array}{ccccc}  1.00 & 0.95 &  2.02 & 0.94 &  1.08 \\
  1.00 & 1.02 & 1.79 & 0.98 & 1.00 \\
1.00 & 0.99 & 1.90 &  0.99 & 1.00   \end{array}\right]$   &  \scriptsize 1000  &  $\scriptsize \left[\begin{array}{c}0.0003 \\ 0.04 \\ 0.01  \end{array}\right]$  \\  \hline 
$\rho_L$  & $\scriptsize \left[\begin{array}{ccccc}0.55 & 2.5 & 0.6 & 0.55 & 
0.95  \\
0.55 & 2.5 & 0.6 & 0.55 & 
0.95 \\
0.55 & 2.5 & 0.6 & 0.55 & 
0.95 \end{array}\right]$ &   \scriptsize 10,000 &  $\scriptsize \left[\begin{array}{c}0.39 \\ 0.31 \\ 0.43   \end{array}\right]$   
\\  \hline 
\mbox{No learning} & $\scriptsize \left[\begin{array}{ccccc}0.0 & 0.0 & 0.0 & 1.0 & 1.0  \\
0.0 & 0.0 & 0.0 & 1.0 & 1.0 \\
0.0 & 0.0 & 0.0 & 1.0 & 1.0 \end{array}\right]$ &   0 &  $\scriptsize \left[\begin{array}{c} 55.55 \\ 68.42 \\ 50.19   \end{array}\right]$    \\ \hline 
\end{tabular}
\end{center}

\begin{figure}[h]
    \centering
\includegraphics[width=50mm,scale=0.5]{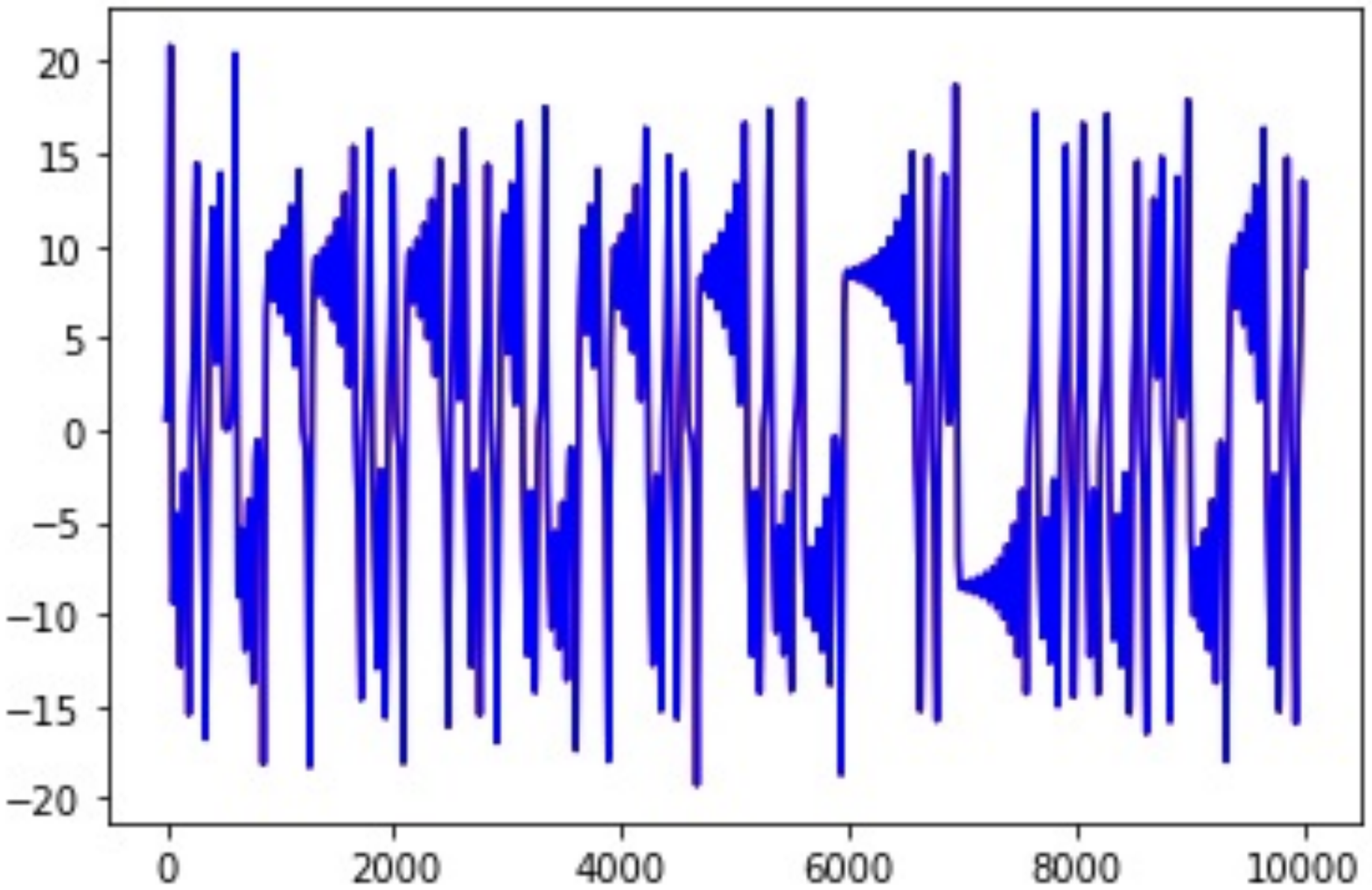}
\includegraphics[width=50mm,scale=0.5]{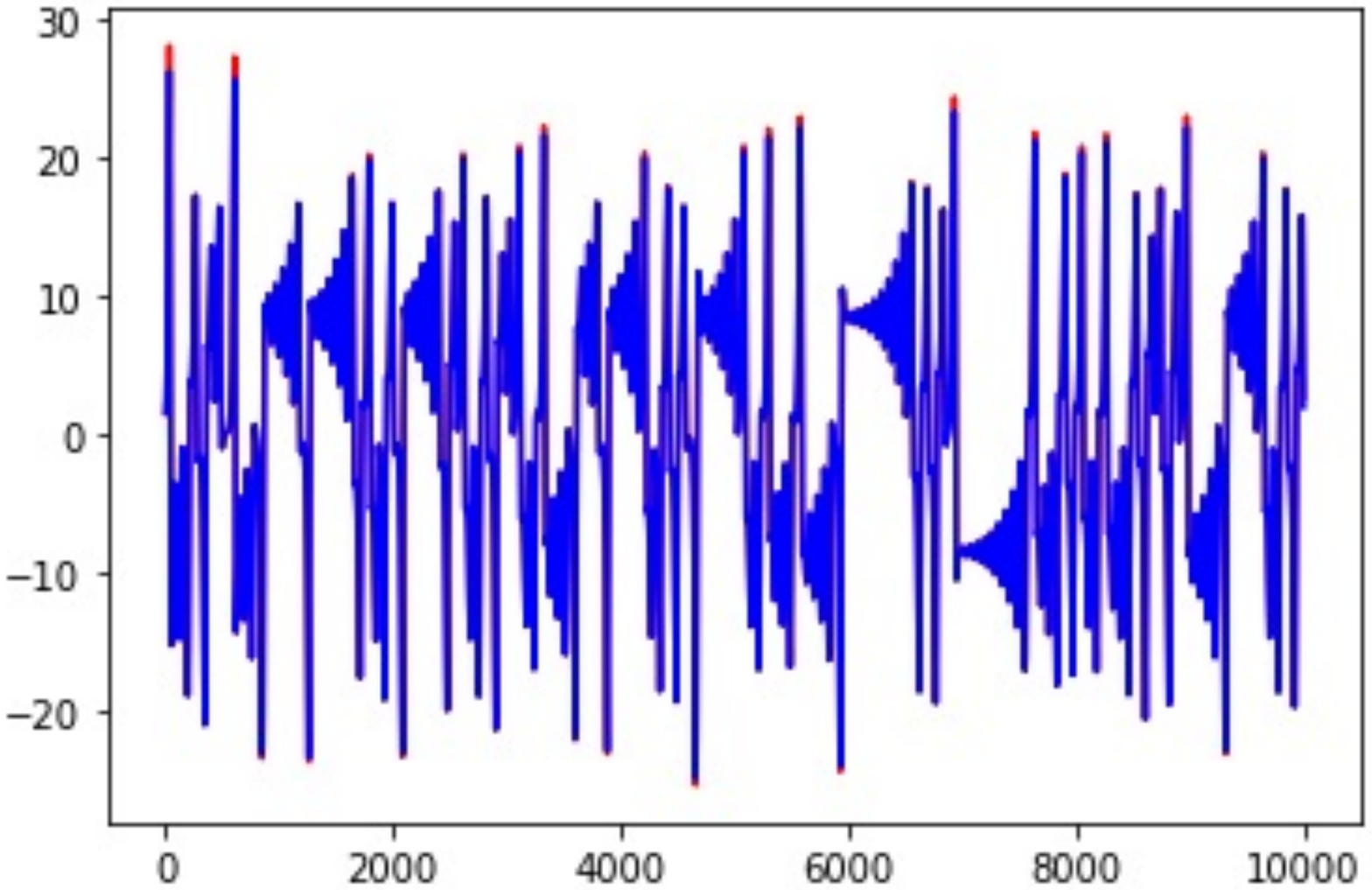}
\includegraphics[width=50mm,scale=0.5]{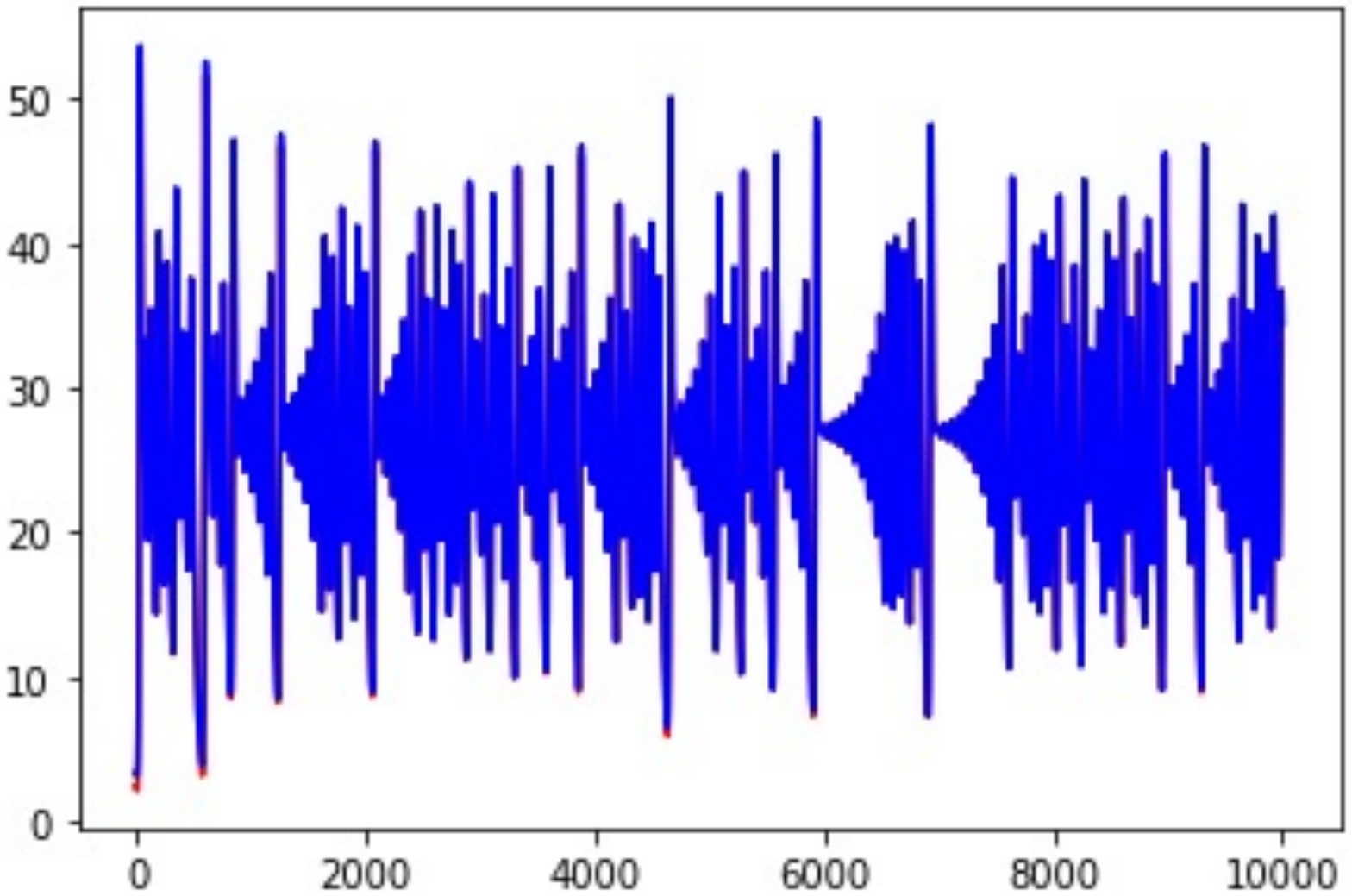}
 \caption{Time series generated by the true dynamics (red) and the approximation with the learned kernel (blue) - x component in the left figure, y component in the middle figure, z component in the right figure.}
 \label{lorentz1}
\end{figure}

\begin{figure}[h]
    \centering
\includegraphics[width=50mm,scale=0.5]{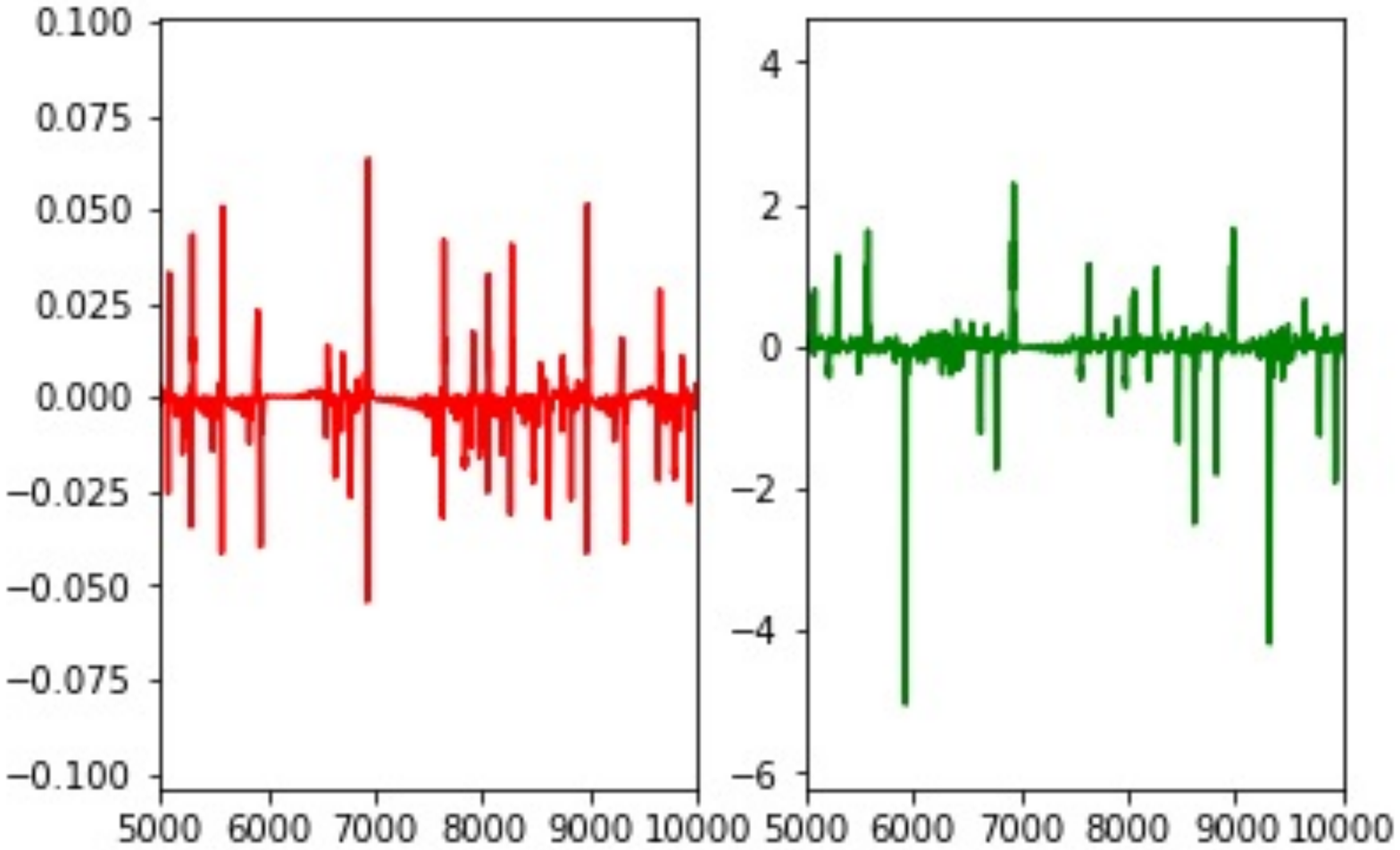}
\includegraphics[width=50mm,scale=0.5]{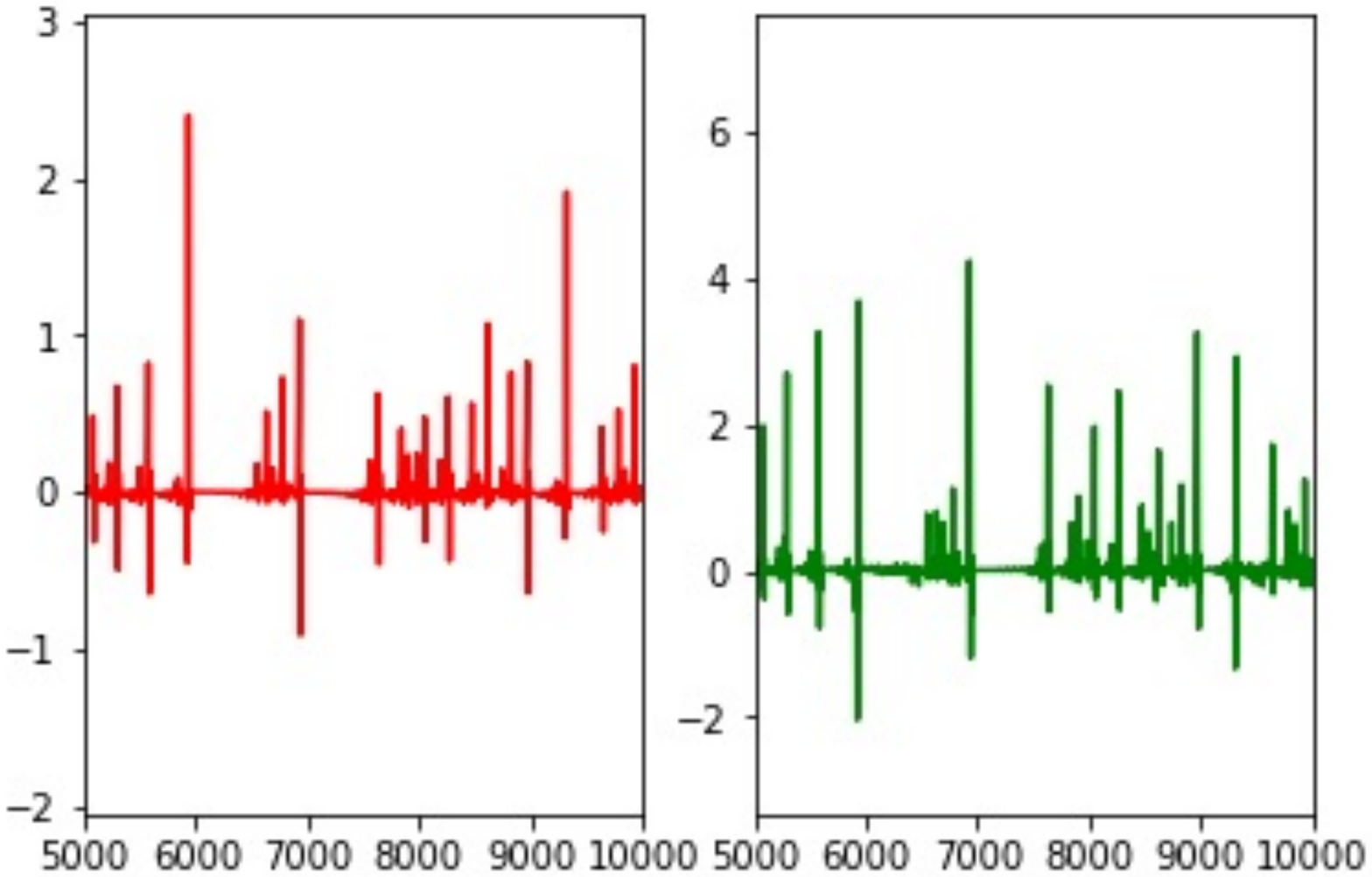}
\includegraphics[width=50mm,scale=0.5]{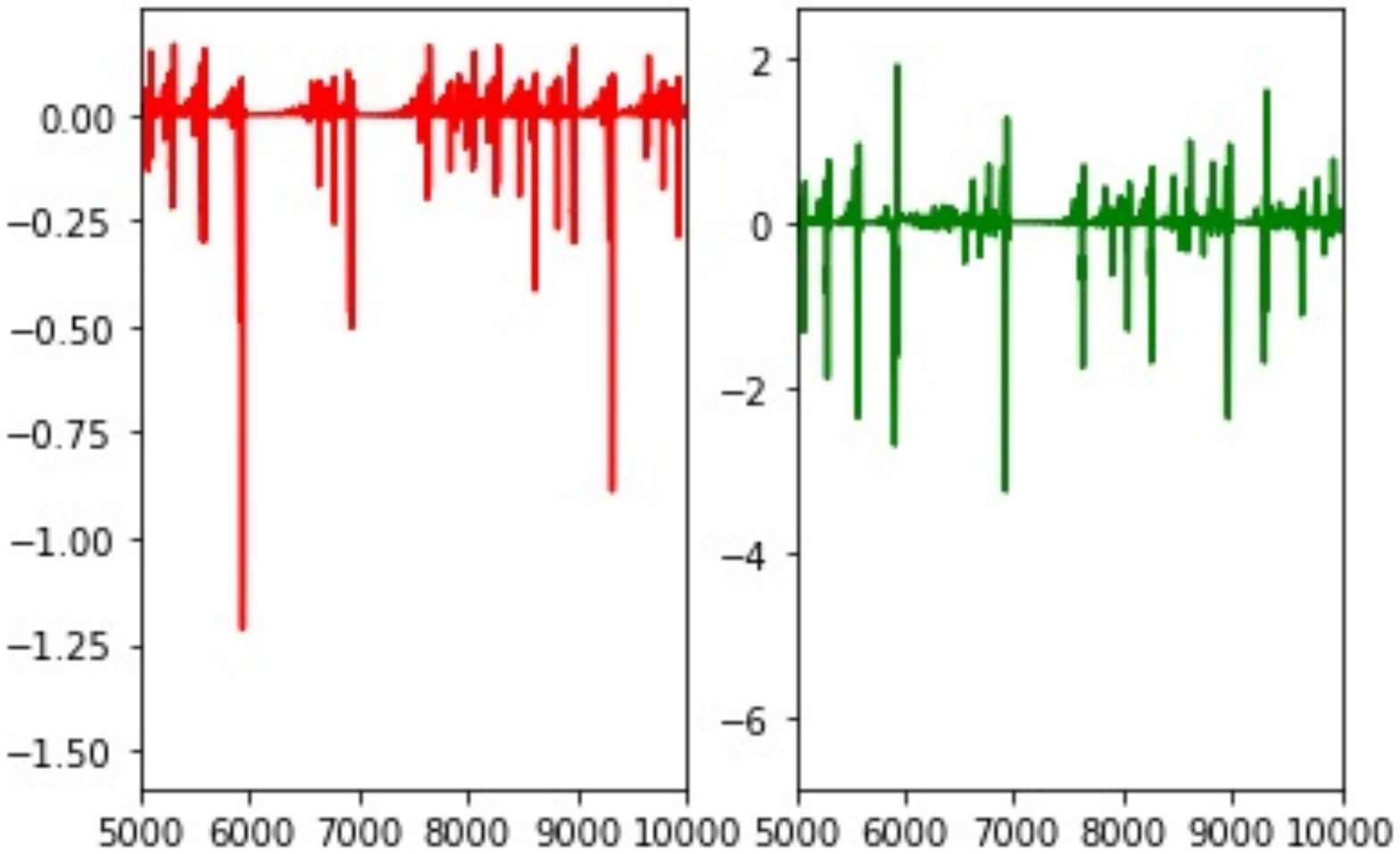}
\caption{Difference between the true and the approximated dynamics with the learned kernel using $\rho$  (red (first, third and fifth from the left)), with the initial kernel (green (second, fourth and sixth from the left)). x-component in the two figures at the left, y-component in the middle two figures, z-component in the right two figures.}
 \label{lorentz2}
\end{figure}

\begin{figure}[h]
\centering
\includegraphics[width=50mm,scale=0.5]{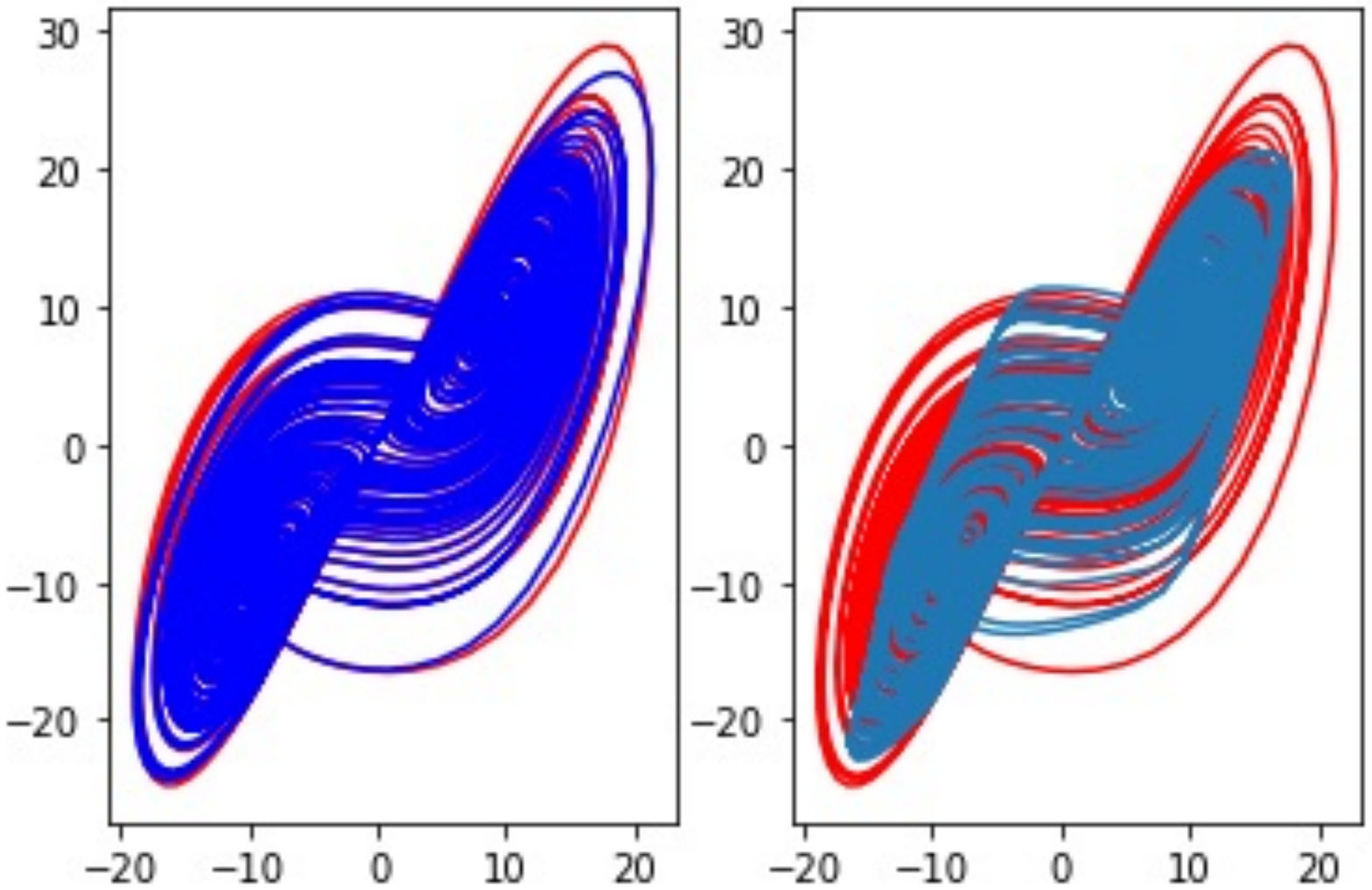}
\includegraphics[width=50mm,scale=0.5]{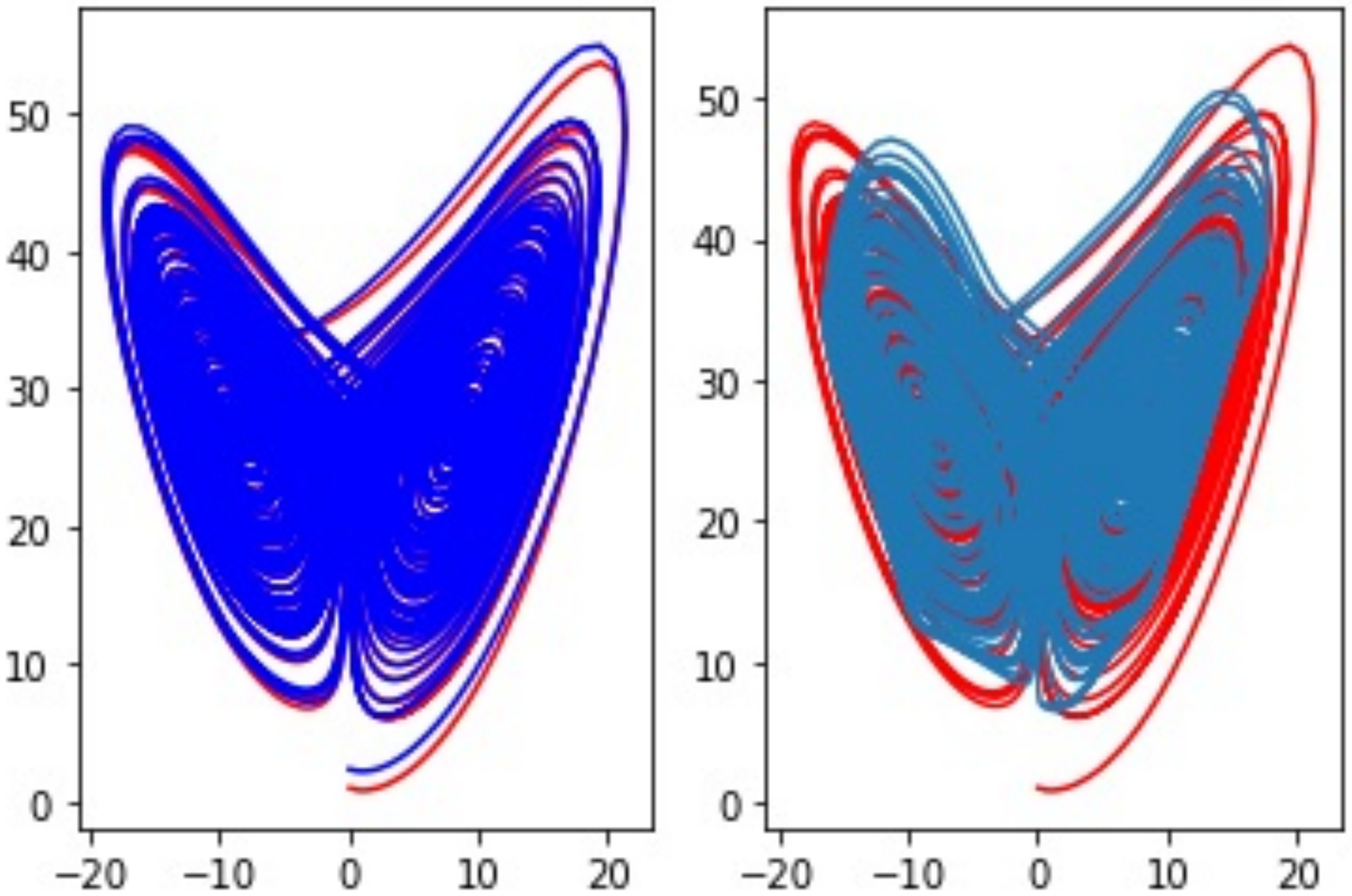}
\includegraphics[width=50mm,scale=0.5]{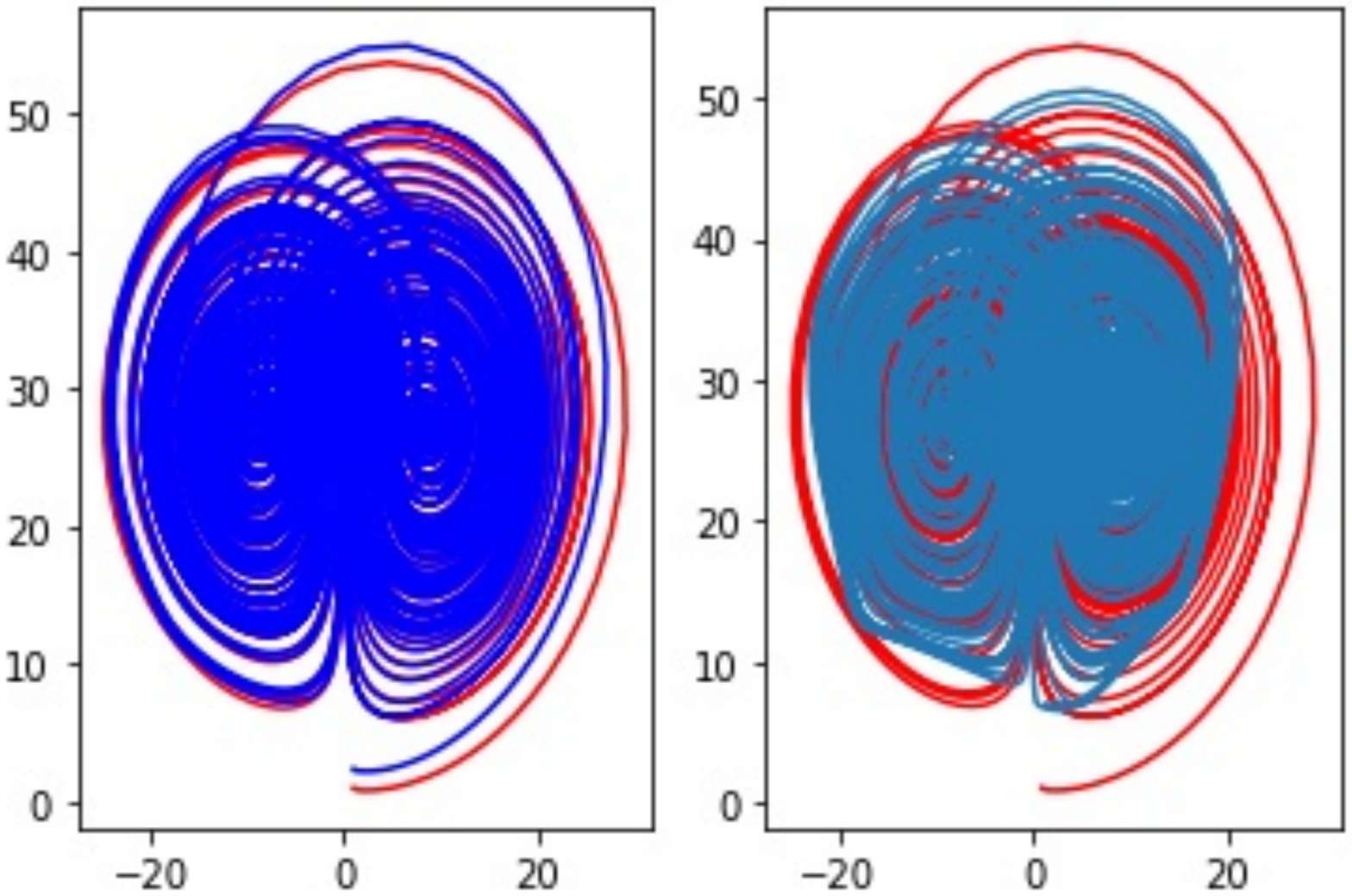}
\caption{Projection of the true attractor and approximation of the attractor using a learned kernel on the XY,XZ and YZ axes  (first, third and fifth from the left), Projection of the true attractor and approximation of the attractor using with initial kernel on the XY,XZ and YZ axes (second, fourth and sixth from the left)}
 \label{lorentz3}
\end{figure}

\begin{figure}[h]
\centering
\includegraphics[width=50mm,scale=0.5]{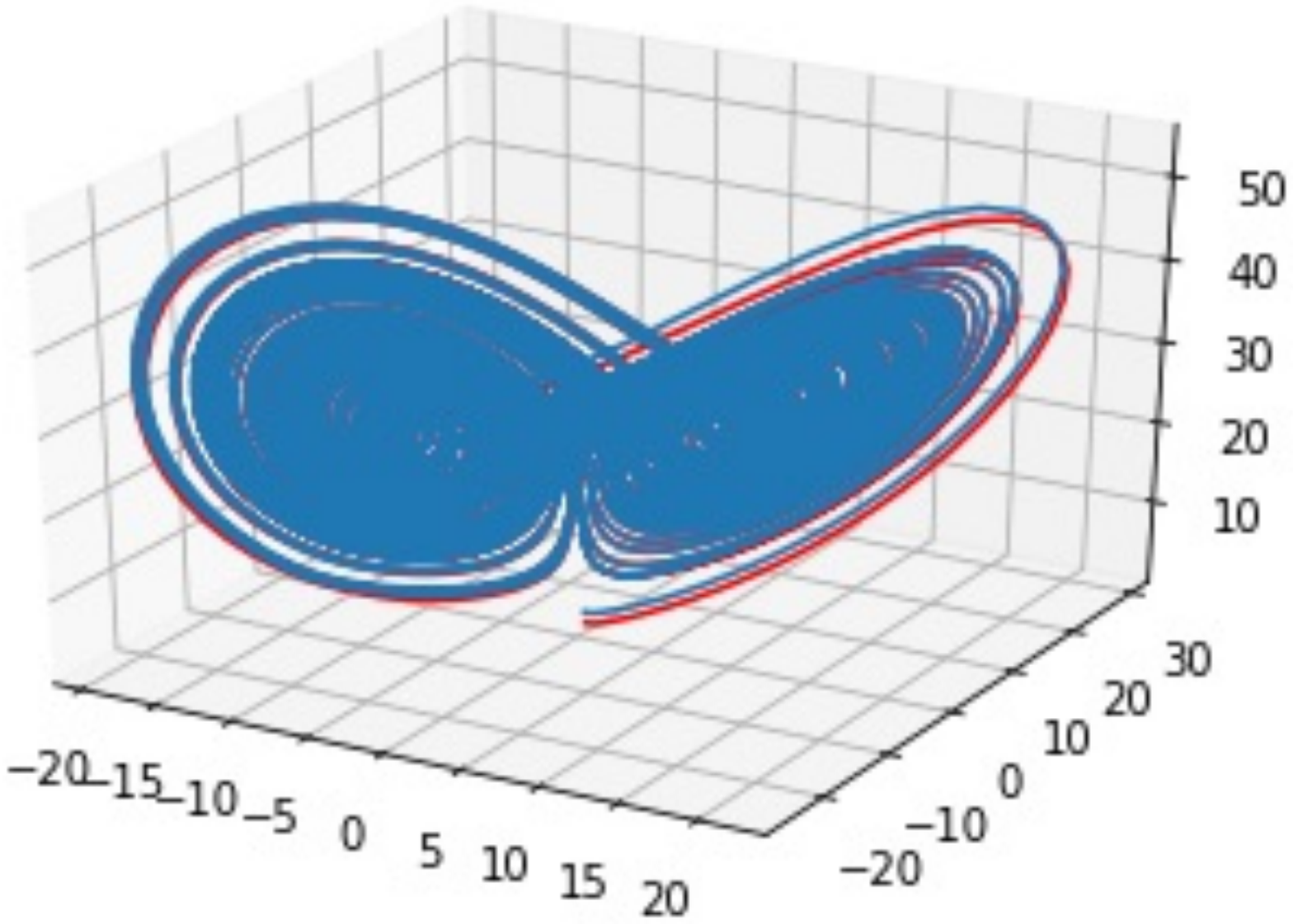}
\includegraphics[width=50mm,scale=0.5]{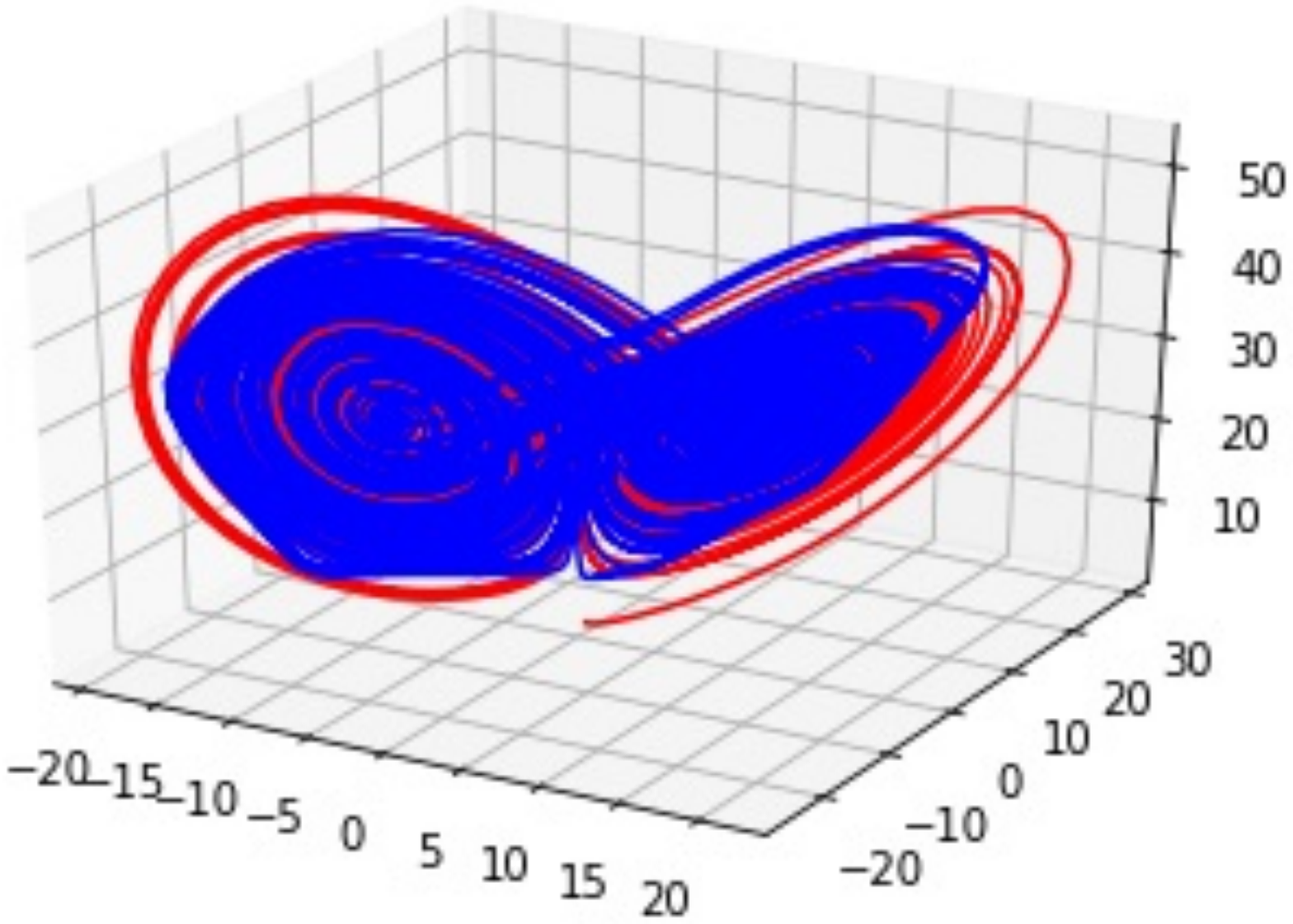}
\caption{True attractor (blue) and approximation of the attractor using a learned kernel (red)  [left], True attractor (blue) and approximation of the attractor using initial kernel (red)  [right]}
 \label{lorentz4}
\end{figure}

Figure \ref{lorentz1} shows the results for an initial condition $(x(0),y(0),z(0))= (0.5, 1.5, 2.5)$ and 10,000 points.  Figure \ref{lorentz2} shows the prediction errors for the case of an approximation with a learned kernel and a kernel without learning.  Figure \ref{lorentz3} shows the projection of the  attractor and its approximation with a learned kernel and a kernel without learning. Figure \ref{lorentz4} shows the attractor with a learned kernel and a kernel without learning.

We also consider a parameterized family of kernels of the form    
\begin{equation}
K_i(x,y)= 
\alpha_{0,i}^2\, \mbox{max}\{0,1-\frac{||x-y||_2^2|}{\sigma_{0,i}}\}+
\alpha_{1,i}^2\, e^{\frac{||x-y||_2^2}{\sigma_{1,i}^2}}+\alpha_2^2 e^{-\frac{||x-y||_2}{\sigma_{2,i}^2}}
+\alpha_{3,i}^2
 e^{- \sigma_{3,i}  \sin^2(\sigma_{4,i} \pi ||x-y||_2^2)}e^{- \frac{||x-y||_2^2}{\sigma_{5,i}^2}}
+\alpha_{4,i}^2  ||x-y||_2^2
\end{equation}
The training and prediction results are shown in the following table with $R_1$ the RMSE  corresponding to 50,000 points with initial conditions $(0.5, 1.5, 2.5)$.

\hspace{-2cm}\begin{center}
\newcolumntype{g}{>{\columncolor{Gray}}c}
\begin{tabular}{ |g |c| c |c| } \hline 
      &   $\scriptsize \left[\begin{array}{ccccccccccc}
      \alpha_{0,1}&\sigma_{0,1}&\alpha_{1,1}&\sigma_{1,1}&\alpha_{2,1}&\sigma_{2,1}&\alpha_{3,1} & \sigma_{3,1} & \sigma_{4,1} & \sigma_{5,1} & \alpha_{4,1}\\
      \alpha_{0,2}&\sigma_{0,2}&\alpha_{1,2}&\sigma_{1,2}&\alpha_{2,2}&\sigma_{2,2}&\alpha_{3,2} & \sigma_{3,2} & \sigma_{4,2} & \sigma_{5,2} & \alpha_{4,2}\\
      \alpha_{0,3}&\sigma_{0,3}&\alpha_{1,3}&\sigma_{1,3}&\alpha_{2,3}&\sigma_{2,3}&\alpha_{3,3} & \sigma_{3,3} & \sigma_{4,3} & \sigma_{5,3} & \alpha_{4,3}       \end{array}\right]$  &   n  &  $R_1$   \\ \hline

   $\rho$ &  $\scriptsize \left[\begin{array}{ccccccccccc}   0.16 &   0.99 &  1.59 &   0.98 &   0.15 &
   0.99 &   0.16 &   1.00 &    1.00 &   0.99&
 -31.28 \\
  -1.03 &    0.99 & -10.96 &   0.10 &  -1.18 &
   0.97 & -1.07 &   1.00 &  1.00 &   0.99 &
  60.87 \\
0.07 & 0.99&  0.68 & 0.89 & 0.07 & 1.00 &
 0.07 & 1.00 & 0.99 & 0.99 & 0.79    \end{array}\right]$   &  \scriptsize 1000  &  $\scriptsize \left[\begin{array}{c}1.0\, 10^{-11} \\ 0.24 \\ 0.17  \end{array}\right]$   \\  \hline 
\mbox{} & $\scriptsize \left[\begin{array}{ccccccccccc}0.0 & 1.0 & 1.0 & 1.0 & 0.0 & 1.0 & 0.0 & 1.0 & 1.0 & 1.0 & 0.0 \\
0.0 & 1.0 & 1.0 & 1.0 & 0.0 & 1.0 & 0.0 & 1.0 & 1.0 & 1.0 & 0.0 \\
0.0 & 1.0 & 1.0 & 1.0 & 0.0 & 1.0 & 0.0 & 1.0 & 1.0 & 1.0 & 0.0 \end{array}\right]$ &   0 &  $\scriptsize \left[\begin{array}{c}54.25 \\ 70.21 \\ 674.92  \end{array}\right]$    \\ \hline 
\end{tabular}
\end{center}


\paragraph{Remarks}
\begin{enumerate}
    \item Convergence results that characterize the error estimates of the difference between a dynamical system and its approximation from data using kernel methods can be found in \cite{BHSIAM2017, lyap_bh}.
    \item In the case of very large datasets, it is possible to reduce the number of points during training by considering greedy techniques as in \cite{gabriel,Sparse_Cholesky}. 
    \item It is possible to include new measurements when approximating the dynamics from data without repeating the learning process. This can be done by working in Newton basis as in \cite{PAZOUKI2011575}. 
    
\end{enumerate}



\section{Conclusion}
Our experiments suggest that using cross-validation (with KF and variants) to learn the kernel used to approximate the vector field of a dynamical system, and thereby its dynamics, significantly improves the accuracy of such approximations. Although our paper is entirely numerical, the simplicity of the proposed approach and the diversity of the experiments raise the question of the existence of a general and fundamental convergence theorem for cross-validation.

\section{Acknowledgment} B.~H. thanks the European Commission for funding through the Marie Curie fellowship STALDYS-792919 (Statistical Learning for Dynamical Systems). H.~O. gratefully acknowledges support by  the Air Force Office of Scientific Research under award number FA9550-18-1-0271 (Games for Computation and Learning). We thank Deniz Ero\u{g}lu, Yoshito Hirata, Jeroen Lamb, Edmilson Roque, Gabriele Santin and Yuzuru Sato for useful comments.

\appendix
\section{Reproducing Kernel Hilbert Spaces}
We give a brief overview of reproducing kernel Hilbert spaces as used in statistical learning
theory ~\cite{CuckerandSmale}. Early work developing
the theory of RKHS was undertaken by N. Aronszajn~\cite{aronszajn50reproducing}.

\begin{definition} Let  ${\mathcal H}$  be a Hilbert space of functions on a set ${\mathcal X}$.
Denote by $\langle f, g \rangle$ the inner product on ${\mathcal H}$   and let $\|f\|= \langle f, f \rangle^{1/2}$
be the norm in ${\mathcal H}$, for $f$ and $g \in {\mathcal H}$. We say that ${\mathcal H}$ is a reproducing kernel
Hilbert space (RKHS) if there exists a function $K:{\mathcal X} \times {\mathcal X} \rightarrow \RR$
such that\\
 i. $K_x:=K(x,\cdot)\in {\cal H}$ for all $x\in {\cal X}$.\\
ii. $K$ spans ${\mathcal H}$: ${\mathcal H}=\overline{\mbox{span}\{K_x~|~x \in {\mathcal X}\}}$.\\
 iii. $K$ has the {\em reproducing property}:
$\forall f \in {\mathcal H}$, $f(x)=\langle f,K_x \rangle$.\\
$K$ will be called a reproducing kernel of ${\mathcal H}$. ${\mathcal H}_K$  will denote the RKHS ${\mathcal H}$
with reproducing kernel  $K$ where it is convenient to explicitly note this dependence.
\end{definition}

The important properties of reproducing kernels are summarized in the following proposition.
\begin{proposition}\label{prop1} If $K$ is a reproducing kernel of a Hilbert space ${\mathcal H}$, then\\
i. $K(x,y)$ is unique.\\
ii.  $\forall x,y \in {\mathcal X}$, $K(x,y)=K(y,x)$ (symmetry).\\
iii. $\sum_{i,j=1}^q\alpha_i\alpha_jK(x_i,x_j) \ge 0$ for $\alpha_i \in \RR$, $x_i \in {\mathcal X}$ and $q\in\mathbb{N}_+$
(positive definiteness).\\
iv. $\langle K(x,\cdot),K(y,\cdot) \rangle=K(x,y)$.
\end{proposition}
Common examples of reproducing kernels defined on a compact domain $\cal{X} \subset\RR^n$ are the 
(1) constant kernel: $K(x,y)= k > 0$
(2) linear kernel: $K(x,y)=x\cdot y$
(3) polynomial kernel: $K(x,y)=(1+x\cdot y)^d$ for $d \in \N_+$
(4) Laplace kernel: $K(x,y)=e^{-||x-y||_2/\sigma^2}$, with $\sigma >0$
(5)  Gaussian kernel: $K(x,y)=e^{-||x-y||^2_2/\sigma^2}$, with $\sigma >0$
(6) triangular kernel: $K(x,y)=\max \{0,1-\frac{||x-y||_2^2}{\sigma} \}$, with $\sigma >0$.
(7) locally periodic kernel: $K(x,y)=\sigma^2 e^{-2 \frac{ \sin^2(\pi ||x-y||_2/p)}{\ell^2}}e^{-\frac{||x-y||_2^2}{2 \ell^2}}$, with $\sigma, \ell, p >0$.

\begin{theorem} \label{thm1}
Let $K:{\mathcal X} \times {\mathcal X} \rightarrow \RR$ be a symmetric and positive definite function. Then there
exists a Hilbert space of functions ${\mathcal H}$ defined on ${\mathcal X}$   admitting $K$ as a reproducing Kernel.
Conversely, let  ${\mathcal H}$ be a Hilbert space of functions $f: {\mathcal X} \rightarrow \RR$ satisfying
$\forall x \in {\mathcal X}, \exists \kappa_x>0,$ such that $|f(x)| \le \kappa_x \|f\|_{\mathcal H},
\quad \forall f \in {\mathcal H}. $
Then ${\mathcal H}$ has a reproducing kernel $K$.
\end{theorem}


\begin{theorem}\label{thm4}
 Let $K(x,y)$ be a positive definite kernel on a compact domain or a manifold $X$. Then there exists a Hilbert
space $\mathcal{F}$  and a function $\Phi: X \rightarrow \mathcal{F}$ such that
$$K(x,y)= \langle \Phi(x), \Phi(y) \rangle_{\cal F} \quad \mbox{for} \quad x,y \in X.$$
 $\Phi$ is called a feature map, and $\mathcal{F}$ a feature space\footnote{The dimension of the feature space can be infinite, for example in the case of the Gaussian kernel.}.
\end{theorem}

\subsection{Function Approximation in RKHSs: An Optimal Recovery Viewpoint} 
In this section we review function approximation in RKHSs from the point of view of optimal recovery as discussed in \cite{owhadi_scovel_2019}. 

\paragraph{Problem {\bf P}:} Given input/output data $(x_1, y_1),\cdots , (x_N , y_N ) \in \mathcal{X} \times \mathbb{R}$,  recover an unknown function $u^{\ast}$ mapping $\mathcal{X}$ to $\mathbb{R}$ such that
$u^{\ast}(x_i)=y_i$ for $i \in \{1,...,N\}$.

In the setting of optimal recovery \cite{owhadi_scovel_2019}  Problem {\bf P} can be turned into a well posed problem by restricting candidates for $u$ to belong to a Banach space of functions $\mathcal{B}$ endowed with a norm $||\cdot||$ and identifying the optimal recovery as the minimizer of the relative error

\begin{equation} \label{game}
    \mbox{min}_v\mbox{max}_u \frac{||u-v||^2}{||u||^2}, 
\end{equation} 
where the max is taken over $u \in \mathcal{B}$ and the min is taken over candidates in $v \in \mathcal{B}$ such that $v(x_i)=u(x_i)=y_i$. For the validity of the constraints $u(x_i) = y_i$,  $\mathcal{B}^{\ast}$, the dual space of $\mathcal{B}$, must contain delta Dirac functions $\phi_i(\cdot)=\delta(\cdot-x_i)$. This problem can be stated as a game between Players I and II and can then be represented as
  \begin{equation}\label{eqdkjdhkjhffORgameban}
\text{\xymatrixcolsep{0pc}\xymatrix{
\text{(Player I)} & u\ar[dr]_{\max}\in \mathcal{B}    &      &v\ar[ld]^{\min}\in L(\Phi,\mathcal{B}) &\text{(Player II)}\\
&&\frac{\|u-v(u)\|}{\|u\|}\,.& &
}}\,
\end{equation}

If $||\cdot||$ is quadratic, i.e. $||u||^2=[Q^{-1}u,u] $ where $[\phi, u]$ stands for the duality product between $\phi \in \mathcal{B}^{\ast}$ and $u \in \mathcal{B}$ and $Q : \mathcal{B}^{\ast}\rightarrow \mathcal{B}$ is a positive symmetric linear bijection (i.e. such that $[\phi, Q \phi] \ge  0$ and $[\psi, Q \phi ] = [\phi, Q \psi]$ for $\phi,\psi \in \mathcal{B}^{\ast} $). In that case the optimal solution of (\ref{game}) has the explicit form 
\begin{equation}\label{sol_rep}v^{\ast}=\sum_{i,j=1}^{N}u(x_i) A_{i,j} Q \phi_j, \end{equation}
where   $A=\Theta^{-1}$ and $\Theta \in \RR^{N \times N}$ is a Gram matrix with entries $\Theta_{i,j}=[\phi_i,Q\phi_j]$.

To recover the classical representer theorem, one defines the reproducing kernel $K$ as $$K(x,y)=[\delta(\cdot-x),Q\delta(\cdot-y)]$$ 
In this case, $(\mathcal{B},||\cdot ||)$ can be seen as an RKHS endowed with the norm
$$||u||^2=\mbox{sup}_{\phi \in \mathcal{B}^\ast}\frac{(\int \phi(x) u(x) dx)^2}{(\int \phi(x) K(x,y) \phi(y) dx dy)}$$
and (\ref{sol_rep}) corresponds to the classical representer theorem 
\begin{equation}\label{eqkjelkjefffhb}
v^{\ast}(\cdot) = y^T AK(x,\cdot),
\end{equation} 
 using the vectorial notation $y^T AK(x,\cdot)=\sum_{i,j=1}^{N}y_iA_{i,j}K(x_j,\cdot)$ with $y_i=u(x_i)$, $A=\Theta^{-1}$ and $\Theta_{i,j} =K(x_i,x_j)$.
  
 Now, let us consider the problem of learning the kernel from data. As introduced in \cite{Owhadi19}, the method of KFs is based on the premise that \emph{a kernel is good if there is no significant loss in accuracy in the prediction error if the number of data points is halved}. This led to the introduction of 
 \[\rho=\frac{||v^{\ast}-v^{s} ||^2}{||v^{\ast} ||^2} \]
  which is the relative error between 
  $v^\ast$, the optimal recovery \eqref{eqkjelkjefffhb} of $u^\ast$ based on the full dataset
  $X=\{(x_1,y_1),\ldots,(x_N,y_N)\}$, and
  $v^s$  the optimal recovery  of both $u^\ast$ and $v^\ast$ based on half of the dataset $ X^s=\{(x_i,y_i)\mid i \in \mathcal{S}\}$ ($\operatorname{Card}(\mathcal{S})=N/2$) which admits the representation
  \begin{equation}
v^s=(y^s)^T A^s K(x^s,\cdot)
  \end{equation}
 with $y^s=\{y_i\mid i \in \mathcal{S}\}$,
 $x^s=\{x_i\mid i \in \mathcal{S}\}$,
 $A^s=(\Theta^s)^{-1}$, $\Theta^s_{i,j}=K(x_i^s,x_j^s)$.
 This quantity  $\rho$ is directly related to the game in (\ref{eqdkjdhkjhffORgameban}) where one is minimizing the relative error of $v^{\ast}$ versus $v^s$. 
Instead of using the entire the dataset $X$ one may use random subsets $X^{s_1}$ (of $X$) for $v^{\ast}$ and random subsets $ X^{s_2}$ (of $X^{s_1}$) for $v^s$. 

Replacing $\|u^\ast \|_\Hc$ by the RKHS norm of the interpolant  of $v^\ast$ (with both testing and training points) in  \eqref{eqkejbdkddbs}   gives an error interval for $v^\ast(x)$ in (\ref{eqkjelkjefffhb}) as
\begin{equation}\label{error_estimate} v^\ast(x) \pm \Delta(v^\ast(x)),\end{equation}
with
\begin{equation}\label{delta} \Delta(v^\ast(x))=\sigma(x) \sqrt{Y^{f,T} K(X^f,X^f)^{-1} Y^f} ,\end{equation}
and where $(X^f,Y^f)$ corresponds to the concatenation of the training and testing points. Local error estimates such as (\ref{error_estimate}) are
classical in Kriging \cite{Wu92localerror} (see also \cite{owhadi2015bayesian}[Thm. 5.1] for applications to PDEs). .

 \subsection{The Maximum Mean Discrepancy} Let ${\cal P}$ be the set of Borel probability measures on ${\cal X}$. Given a probability distribution $P$ we define its
   kernel mean embedding (with respect to a kernel $k$ with RKHS $\Hc$) as 
$$\begin{array}{rcl} \mu_P: {\cal P}& \rightarrow & {\cal H}  \\
P & \mapsto& \int_{\cal X} k(x,y) dP(y) =: \mu_k(P)
\end{array}
$$
The maximum mean discrepancy (MMD) between two probability  measures $P$ and $Q$ is then defined as the distance between two such embeddings and can be expressed as
$$\begin{array}{rcl} \mbox{MMD}(P,Q) &:=& ||\mu_P-\mu_Q||_{\cal H}, \\ 
&=& \big(\mathbb{E} _{x,{x}^{\prime}}(k(x,x^{\prime}))+\mathbb{E} _{y,{y}^{\prime}}(k(y,y^{\prime}))-2 \mathbb{E} _{x,y}(k(x,y)\big)^{\frac{1}{2}}
\end{array}$$
where $x$ and $x^{\prime}$ are independent random variables drawn according to $P$, $y$ and $y^{\prime}$ are independent random variables drawn according to $Q$, and $x$ is independent of $y$.

Given i.i.d. samples from $X:=\{x_1,...,x_m\}$ and $Y:=\{y_1,...,y_n\}$, from $P$ and $Q$ respectively, recall that
the MMD in RKHSs is defined as the difference between the kernel mean embeddings defined as as follows.  
Given  i.i.d samples $(x_1,\cdots,x_m)$ from $P$ and $(y_1,\cdots,y_n)$ from $Q$, the MMD between 
the empirical distributions $(\delta_{x_1}+\cdots+\delta_{x_m})/m$ and
$(\delta_{y_1}+\cdots+\delta_{y_n})/n$ is an unbiased estimate of $\mbox{MMD}(P,Q)$ with the representation
\begin{equation}\label{eqlkedkejd}
\mbox{MMD}_u^2 := \frac{1}{m^2}\sum_{i, j=1}^m k(x_i,x_j)+\frac{1}{n^2}\sum_{i, j=1}^n k(y_i,y_j)-\frac{2}{n m}\sum_{i=1}^m \sum_{j=1}^n k(x_i,y_j)
\end{equation}

 {\small{}\bibliographystyle{unsrturl}
\bibliography{kernel_flows_for_koopman}
}{\small\par}
\end{document}